\documentclass[10pt,twocolumn,letterpaper]{article}

\usepackage{wacv}              

\newcommand*{\ourmodel}{DVPSFormer\@\xspace}

\usepackage[utf8]{inputenc} 
\usepackage[T1]{fontenc}    
\usepackage{url}            
\usepackage{booktabs}       
\usepackage{amsmath}
\usepackage{amsfonts}       
\usepackage{nicefrac}       
\usepackage{microtype}      
\usepackage{xcolor}         
\usepackage[table]{xcolor}

\definecolor{wacvblue}{rgb}{0.21,0.49,0.74}
\usepackage[pagebackref,breaklinks,colorlinks,allcolors=wacvblue]{hyperref}

\input{abbrv}

\usepackage{newtxtext}
\usepackage{acro}
\usepackage{amsmath}
\usepackage{graphicx}
\usepackage{booktabs}
\usepackage{multirow}
\usepackage{wrapfig}
\usepackage{lipsum}
\usepackage{upgreek}

\usepackage[capitalize]{cleveref}
\crefname{section}{Sec.}{Secs.}
\Crefname{section}{Section}{Sections}
\Crefname{table}{Table}{Tables}
\crefname{table}{Tab.}{Tabs.}
\Crefname{algorithm}{Algorithm}{Algorithms}
\crefname{algorithm}{Alg.}{Algs.}

\def\wacvPaperID{58} 
\def\confName{WACV}
\def\confYear{2027}

\title{
    \ourmodel: Efficient Online Depth-aware Video Panoptic Segmentation for Autonomous Driving
}

\author{
Yung-Hsu Yang\textsuperscript{1} \quad Luigi Piccinelli\textsuperscript{1} \quad Siyuan Li\textsuperscript{1} \quad Mattia Segu\textsuperscript{1} \quad Lei Ke\textsuperscript{1} \quad Martin Danelljan\textsuperscript{1} \\[0.15cm] Yuqian Fu\textsuperscript{2} \quad Zuria Bauer\textsuperscript{1} \quad Fisher Yu\textsuperscript{1} \quad Hermann Blum\textsuperscript{3} \quad Marc Pollefeys\textsuperscript{1,4} \\[0.3cm]
\small $^1$ETH Z\"urich \quad $^2$INSAIT, Sofia University \quad $^3$University of Bonn \quad $^4$Microsoft \\
}

\begin{document}
\maketitle

\begin{abstract}
Safe autonomous navigation requires a holistic understanding of dynamic environments, necessitating the simultaneous estimation of metric depth, semantic segmentation, and instance trajectories.
While depth-aware video panoptic segmentation (DVPS) unifies these tasks, existing approaches often rely on computationally expensive, multi-stage pipelines or offline tracking, rendering them unsuitable for real-time decision-making.
To address this, we propose \ourmodel, a unified online architecture designed for efficient 4D scene understanding.
Central to our approach is explicit scene discretization (ESD), a novel mechanism that leverages segmentation queries to represent foreground and background regions, enabling a discrete-to-continuous (D2C) depth head to decode metric depth in a single pass.
This tightly couples semantic and geometric learning while significantly reducing latency.
Furthermore, we propose an online majority voting (OMV) mechanism that exploits temporal consistency to refine classification during instance tracking.
\ourmodel establishes a new state-of-the-art on the Cityscapes-DVPS and SemKITTI-DVPS benchmarks, offering a streamlined solution for online robotic perception.
Code and models are available at \href{https://royyang0714.github.io/DVPSFormer}{royyang0714.github.io/DVPSFormer}.
\end{abstract}

\section{Introduction}
\label{sec:intro}

For autonomous agents operating in complex, dynamic urban environments, a holistic 4D scene understanding is crucial for machine perception.
To navigate safely, an autonomous vehicle must simultaneously identify objects and background, \ie panoptic segmentation (PS)~\cite{kirillov2019panoptic,vodisch2024good}, localize them in metric 3D space, \ie metric monocular depth estimation (MMDE)~\cite{eigen2014depth}, and predict the trajectories of objects over time, \ie instance tracking~\cite{yang2019video,fantrack,eagermot}.
This convergence of tasks is formalized as depth-aware video panoptic segmentation (DVPS), a critical capability for downstream applications in robotics~\cite{zhou2019does, wofk2019fastdepth} and self-driving~\cite{kim2025semantic, autonomous, brodermann2025cafuser, broedermannn2025dgfusion, athar20234d}.

\begin{figure}[t]
    \centering
    \includegraphics[width=1.0\linewidth]{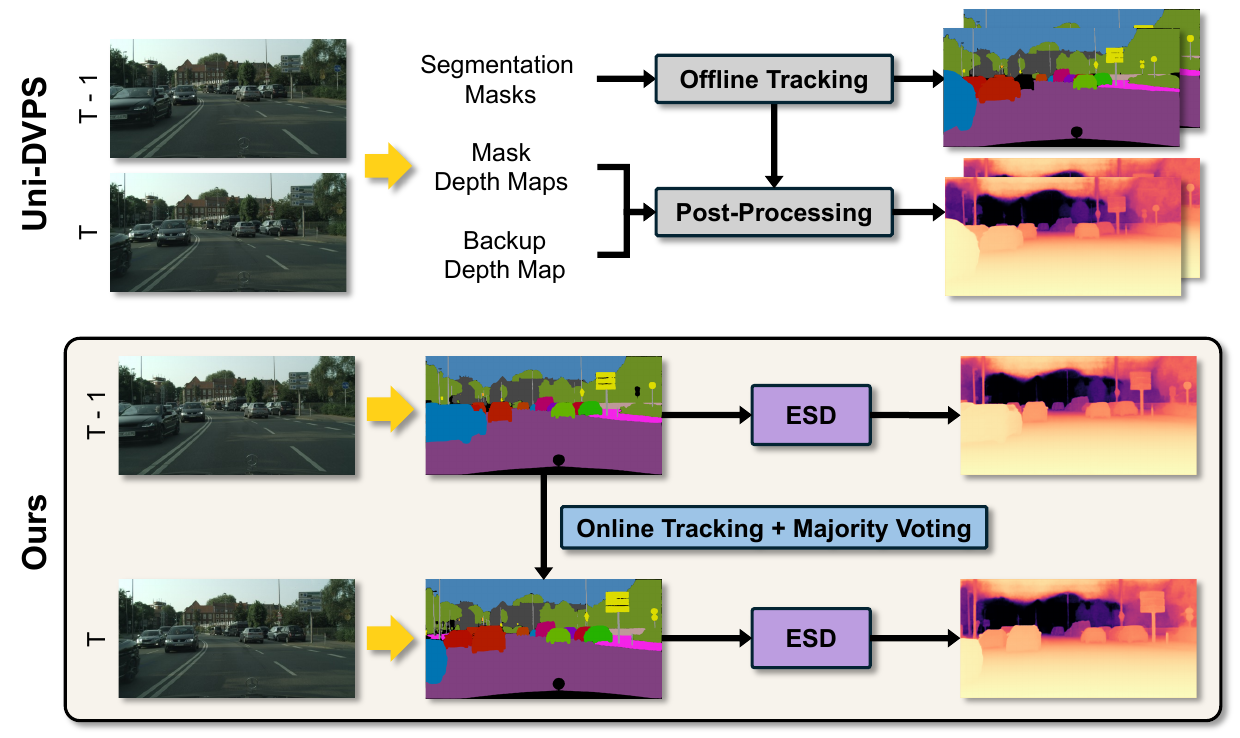}
    \vspace{-15pt}
    \caption{
        \textbf{Intuition on \ourmodel.}
        Compared to the previous state-of-the-art method~\cite{jiyeon2024unidvps}, \ourmodel is an \textit{efficient} and \textit{online} architecture for autonomous driving.
        We treat the segmentation pipeline as the explicit scene discretization process to estimate metric depth in one pass, and design the online tracking pipeline with majority voting to obtain better segmentation.
    }
    \vspace{-5pt}
    \label{fig:teaser}
\end{figure}

However, the main challenge of deploying DVPS on autonomous vehicles is the latency and ``online'' operational requirements.
While previous approaches~\cite{qiao2021vip, yuan2022polyphonicformer, petrovai2023monodvps, jiyeon2024unidvps} have shown promising results, they overlook the trade-off between performance and efficiency due to their convoluted pipelines.
In particular, DVPS can be divided into two sub-tasks, such as depth-aware panoptic segmentation (DPS)~\cite{gao2022panopticdepth, wu2024efficientdps}, which is the combination of PS and MMDE, and instance tracking.
For DPS, previous methods~\cite{qiao2021vip, yuan2022polyphonicformer, petrovai2023monodvps} design separate task-specific modules, thus missing the opportunity to fully exchange information between segmentation and depth.
Despite Uni-DVPS~\cite{jiyeon2024unidvps} proposing a unified transformer architecture~\cite{vaswani2017attention, cheng2022masked} for DPS, its convoluted depth estimation pipeline, as shown in \cref{fig:teaser}, still complicates the pipeline while yielding suboptimal inference speed and results.
For instance tracking, previous methods~\cite{qiao2021vip, yuan2022polyphonicformer, petrovai2023monodvps} either rely on post-processing and extra feature extraction or \textit{offline} tracking~\cite{jiyeon2024unidvps}, which requires access to future video frames, making them unsuitable for real-world applications.

To address the aforementioned challenges, we focus on addressing \textit{online} DVPS for autonomous driving and propose \ourmodel, a unified and efficient architecture.
For DPS, we re-frame the panoptic segmentation branch as the explicit scene discretization (ESD) process to better unify PS and MMDE tasks.
Unlike using slot-attention~\cite{locatello2020object} to perform internal scene discretization for MMDE~\cite{piccinelli2023idisc}, we obtain the foreground and background information from the segmentation \textit{explicitly}.
We design the depth estimation head as the discrete-to-continuous (D2C) transformation to decode metric depth in a single pass.
This approach tightly couples semantic and geometric learning, allowing depth supervision to directly refine object segmentation while reducing computational overhead.

For instance tracking, to compare the performance with \textit{offline} method~\cite{jiyeon2024unidvps}, we propose a lightweight tracking head to leverage the unified queries for association and enhance the features with popular quasi-dense similarity training~\cite{pang2021quasi,fischer2023qdtrack, hu2022monocular, fischer2022cc, wu2022defense}.
Moreover, we propose an \textit{online} majority voting (OMV) to account for misclassifications in the temporal association and correct them through majority vote.
This allows \ourmodel to use temporal information to achieve better overall performance while still being run in an online manner for autonomous driving scenes.
Meanwhile, we also address the specific data challenges of the autonomous driving domain.
Driving datasets often contain sparse or noisy ground truth derived from projected LiDAR.
To mitigate this, we adjust the point-based supervision~\cite{cheng2022masked} algorithm to filter artifacts during training.
This leads to better panoptic quality with efficient training.

We evaluate \ourmodel on the two main DVPS benchmarks, Cityscapes-DVPS~\cite {cordts2016cityscapes} and SemKITTI-DVPS~\cite{behley2019semantickitti}.
Our model consistently outperforms the previous state-of-the-art (SOTA) methods, while achieving higher inference speed and improved efficacy.
Compared to previous SOTA for running $20$ frames association, \ourmodel achieve about $18 \times$ faster inference speed, demonstrating the value of our \textit{online} approach for autonomous driving agents.
To conclude, we propose a novel depth estimation approach for DVPS, which treats the segmentation process as an explicit scene discretization, simplifying the overall pipeline while achieving better segmentation and depth results.
Furthermore, we design the online tracking pipeline with an online majority voting mechanism to enhance applicability and yield better performance.

\section{Related Work}
\label{sec:rel}

\noindent{}\textbf{Panoptic Segmentation (PS)}~\cite{kirillov2019panoptic} requires performing pixel-level semantic and instance segmentation jointly.
The problem is typically tackled with two heads~\cite{kirillov2019panoptic, xiong2019upsnet, cheng2020panoptic}, which generate both pixel-level classification and instance recognition respectively and are merged in the final panoptic segmentation mask via post-processing~\cite{kirillov2019panoptic}.
More recently, architectures have been designed to tackle the two tasks in a unified manner.
Dynamic kernel designs~\cite{li2021fully, zhang2021k} encode things and stuff into specific kernels, and unify segmentation and instance mask predictions in a single architecture.
Other works~\cite{cheng2021per, cheng2022masked, yu2022k} have focused on encoding things and stuff as a set of unified representations based on the popular DETR-like~\cite{locatello2020object, zhu2020deformable} architecture, namely transformer queries~\cite{vaswani2017attention}.
Our method uses Mask2Former~\cite{cheng2022masked} as the segmentation model and treats this process as explicit scene discretization to decode final depth estimation.
This design effectively integrates panoptic segmentation, depth estimation, and instance tracking within a single framework.
We further leverage the proposed online majority voting to enhance the panoptic quality using temporal information.

\noindent{}\textbf{Metric Monocular Depth Estimation (MMDE)}~\cite{eigen2014depth} is essential for 3D world understanding from a single image.
MMDE has seen a recent surge in applicability thanks to foundation models~\cite{piccinelli2024unidepth, piccinelli2025unidepthv2, piccinelli2025unik3d, hu2024metric3dv2, bochkovskii2024depthpro} yielding impressive zero-shot performance.
However, most foundation models do not dive into developing an optimal scene representation.
iDisc~\cite{piccinelli2023idisc} utilizes slot-attention~\cite{locatello2020object} to learn the internal scene discretization and decode the depth estimation according to the knowledge of the foreground and background without explicit segmentation guidance.
In contrast, \ourmodel treats the segmentation pipeline as the explicit scene discretization process and estimates the depth in one pass.

\noindent{}\textbf{Instance Tracking.}
To enable temporal understanding for dynamic instances, Video Panoptic Segmentation (VPS)~\cite{kim2020video} combines instance tracking~\cite{yang2019video} with PS.
\ourmodel designs the online tracking pipeline inspired by multiple object tracking (MOT)~\cite{pang2021quasi,hu2022monocular,fischer2022cc,fischer2023qdtrack} to associate instance masks with similarity learning while using a lightweight tracking head to enhance transformer queries.
We further propose online majority voting to enhance segmentation performance, leveraging temporal information through tracking.

\begin{figure*}[t]
    \centering
    \includegraphics[width=0.98\linewidth]{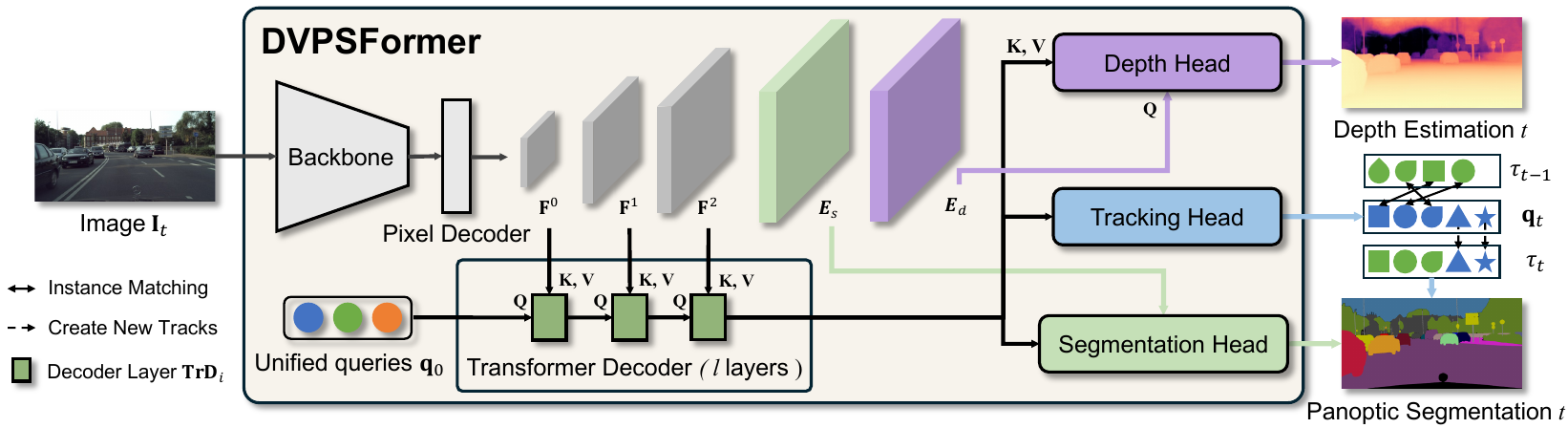}
    \vspace{-5pt}
    \caption{
        \textbf{Model Architecture.}
        \ourmodel is an online unified architecture for DVPS.
        We treat the segmentation pipeline as an explicit scene discretization process and utilize a discrete-to-continuous depth head to decode MMDE and track instances with representative tracking queries, which are trained using similarity learning.
        We plot three transformer decoder layers $\mathrm{TrD}_{i}$ in the figures, where the decoder operates in a round-robin fashion.
    }
    \vspace{-10pt}
    \label{fig:overview}
\end{figure*}

\noindent{}\textbf{Depth-aware Video Panoptic Segmentation (DVPS)} is proposed by ViP-DeepLab~\cite{qiao2021vip}, which consists of three foundational tasks, \ie PS, MMDE, and instance tracking.
ViP-DeepLab extends Panoptic-DeepLab~\cite{cheng2020panoptic} with a depth estimation head and stitches the consecutive input images across time to perform instance tracking as a strong baseline.
PolyphonicFormer~\cite{yuan2022polyphonicformer} extends K-Net~\cite{zhang2021k} with a depth estimation head and proposes query-linking to help depth estimation with segmentation by adding panoptic queries to depth queries.
They convert the instance masks to 2D bounding boxes and use RoI Align~\cite{he2017mask} to extract appearance features for tracking heads~\cite{pang2021quasi, fischer2023qdtrack}.
On the contrary, we propose ESD inspired by iDisc's internal scene discretization, but replace slot-attention with explicit segmentation supervision as a novel paradigm, and our tracking head can directly use the unified query as appearance features without the huge computational overhead.
MonoDVPS~\cite{petrovai2023monodvps} proposes using the pose and optical flow estimation architecture to achieve self-supervised learning for DVPS

Uni-DVPS~\cite{jiyeon2024unidvps} proposes a unified architecture based on Mask2Former~\cite{cheng2022masked} for DPS, and directly uses transformer queries as appearance features to associate instances in an offline manner.
However, the per-mask depth estimation fails to fully leverage the semantic information for depth estimation and propagate the depth supervision to directly aid segmentation.
Moreover, instance queries are not sufficiently representative to distinguish objects across time, and the offline design is not only impractical but also leads to a heavier computational cost for longer sequences.
To address these issues, we propose \ourmodel, which treats the segmentation pipeline as explicit scene discretization, and directly utilizes segmentation queries to generate depth estimation.
Furthermore, we enhance queries through instance contrastive learning~\cite{pang2021quasi, wu2022defense, fischer2023qdtrack} and design an online tracking mechanism based on pure embedding similarity, achieving high efficiency.
Together, we propose an efficient and superior architecture for DVPS.

\section{Method}
\label{sec:method}

We first introduce the overall design of \ourmodel in \cref{sec:method:overview}.
Then, we illustrate how we treat the segmentation process as explicit scene discretization to estimate depth in \cref{sec:method:depth}.
\cref{sec:method:tracking} describes how we extract tracking queries from the unified architecture and our proposed online majority voting.
Finally, the overall training criterion is illustrated in \cref{sec:method:loss}.

\subsection{Overview}
\label{sec:method:overview}
\ourmodel follows the encoder-decoder paradigm~\cite{cheng2022masked,piccinelli2023idisc,jiyeon2024unidvps} with three additional task-specific heads, as shown in \cref{fig:overview}.
We aim to predict segmentation masks $\mathbf{M}$ with per-mask class predictions $\mathbf{C}$ for panoptic segmentation (PS) with metric depth estimation $\mathbf{D}$ simultaneously from a single input image ($\mathbf{I}_t$) at frame $t$.
We track the instances as trajectories $\uptau$ over time in an online manner from the input video sequence.
In the encoder stage, \ourmodel presents a shared backbone and a shared pixel decoder.
We extract the unified multi-scale features $\mathcal{F} = \{\mathbf{F}^0, \mathbf{F}^1, \mathbf{F}^2\}$ from $\mathbf{I}$, corresponding to $1/32$, $1/16$, and $1/8$ resolutions of the input image size.
Due to the difference in output spaces, unlike Uni-DVPS, we use separate convolutional layers to obtain task-specific pixel representations $\mathbf{E}_s$ and $\mathbf{E}_d$ for PS and MMDE, which are $1/4$ of the image resolution.

In the decoder stage, our design involves multiple transformer decoder layers~\cite{vaswani2017attention, carion2020end, zhu2020deformable, cheng2021per, cheng2022masked}, which gradually refine the set of learnable unified queries $\mathbf{q}$ with $\mathcal{F}$ over $l$ layers.
For the $i$-th transformer decoder layer, \ourmodel refines the queries $\mathbf{q}^{i}$ by one cross-attention ($\mathrm{CA}$) conditioned on the set of multi-scale features $\mathcal{F}$, followed by a self-attention ($\mathrm{SA}$) layer and a Multi-Layer Perceptron ($\mathrm{MLP}$).
For the sake of completeness, we define the $i$-th ``transformer decoder layer'', \ie $\mathrm{TrD}_{i}$, as $\mathrm{MLP}_{i}(\mathrm{SA}^{i}(\mathrm{CA}^{i}(\cdot,\cdot)))$ and formulate the refinement process as follows:
\begin{equation}
    \mathbf{q}_{i+1} = \mathrm{TrD}_{i}(\mathbf{q}_{i}, \mathbf{F}^{i\mathrm{mod}3}),
\end{equation}
where ``$i\mathrm{mod}3$'' means $i$ modulo three, and three is the total number of feature scales, representing the round-robin fashion.

We generate segmentation masks $\mathbf{M}_{i}$ using the dot product between $\mathbf{q}_{i}$ processed by $\mathrm{MLP}_{mask}$ and high-resolution pixel representation $\mathbf{E}_{s}$ as
\begin{equation}
    \mathbf{M}_{i} = \begin{cases}
        1, \text{ if } \sigma(\mathrm{MLP}_{mask}(\mathbf{q}_{i}) \cdot \mathbf{E}_{s}) > 0.5 \\
        0, \text{ otherwise} 
    \end{cases},
\end{equation}
where $\sigma$ represents the sigmoid function, \ie $\sigma(x) = \frac{1}{1 + e^{-x}}$.
The masks are also utilized at the $(i+1)$-th layer for mask attention~\cite{cheng2022masked}, thus we pass the refined queries to the segmentation head after each transformer decoder layer, including the initial query, \ie $\mathbf{q}_0$.
Per-mask class predictions $\mathbf{C}_{i}$ are obtained by projecting the segment queries with one linear layer followed by the $\mathrm{softmax}$ function.
After $l$-layer refinement, we pass the unified queries $\mathbf{q}_{l-1}$ to the depth estimation head and tracking head for the other tasks.

\subsection{Explicit Scene Discretization}
\label{sec:method:depth}
One challenge of the unified model for PS and MMDE estimation is to align the multi-task output correctly.
Previous works~\cite{gao2022panopticdepth, yuan2022polyphonicformer,jiyeon2024unidvps} perform per-segment depth estimation and compute loss with the ground truth (GT) depth mask according to the matching results between mask prediction and GT mask.
This design will lead not only to incomplete depth estimation due to the coverage of the mask prediction, but also to the misalignment between training and inference time due to the mask selection based on the masks' scores.
Uni-DVPS~\cite{jiyeon2024unidvps} leverages the extra backup query to generate the complete depth map and combines it with the per-segment depth estimation to overcome the incomplete depth prediction, but further convolutes the depth estimation process and still suffers from train-test misalignment.

\begin{figure}[t]
    \centering
    \includegraphics[width=1.00\linewidth]{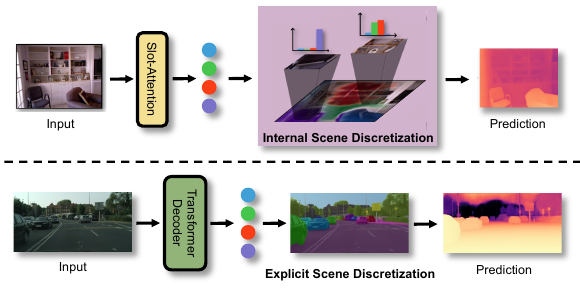}
    \vspace{-15pt}
    \caption{
        \textbf{Scene Discretization.}
        The upper part is the iDisc~\cite{piccinelli2023idisc} approach to obtain the internal scene discretization, while the lower part is ours, which uses the segmentation model to discretize the scene explicitly.
    }
    \vspace{-10pt}
    \label{fig:ESD}
\end{figure}

Inspired by iDisc~\cite{piccinelli2023idisc}, which leverages slot attention~\cite{locatello2020object} to learn the internal scene representation and decode the final depth estimation, we treat the segmentation pipeline as an explicit scene discretization (ESD) process.
As shown in \cref{fig:ESD}, compared to internal scene discretization, the segmentation model~\cite{cheng2022masked} is explicitly trained on the annotated foreground and background GT.
This allows those discrete scene representations, \ie unified queries, to be more representative compared to the slot-attention queries.
To convert these explicit discrete scene representations to continuous depth prediction, \ourmodel uses a simple cross-attention $\mathrm{CA}_{esd}$ between the high-resolution pixel representation $\mathbf{E}_{d}$ and $\mathbf{q}_{l-1}$ processed by $\mathrm{MLP}_{depth}$ as a discrete-to-continuous (D2C) transformation.
To be more specific, the pixel representation $\mathbf{E}_d$ has a shape of (B, C, H, W), while the unified queries tensor has a shape of (B, N, C).
We will reshape $\mathbf{E}_d$ as (B, H $\times$ W, C), and use it as the query while using the unified queries as key and value for cross-attention $\mathrm{CA}_{esd}$.
The cross-attention output tensor is processed by an MLP ($\mathrm{MLP}_{esd}$) to project the tensor channel dimension to one as a log-scale depth prediction.
Finally, we reshape the MLP output tensor back to (B, 1, H, W) as the depth estimation $\mathbf{D}$.
The computation of the depth estimation $\mathbf{D}$ can be formally described as follows:
\begin{equation}
    \mathbf{D} = \exp(\mathrm{MLP}_{esd}(\mathrm{CA}_{esd}(\mathbf{E}_{d}, \mathrm{MLP}_{depth}(\mathbf{q}_{i})))).
\end{equation}

This design allows \ourmodel to directly decode the complete depth map at one pass with unified queries, which not only simplifies the depth estimation head but also successfully aligns the training and inference pipeline without being affected by the mask selection.
Our design also enables the depth loss to directly affect the segmentation queries, further improving the panoptic quality (PQ).
Moreover, our design only requires the output of the last transformer decoder layer, \ie $\mathbf{q}_{l-1}$, because we only need the final scene representation to estimate the depth, which saves training resources compared to per-layer supervision.

\begin{figure}[t]
    \centering
    \vspace{5pt}
    \includegraphics[width=1.0\linewidth]{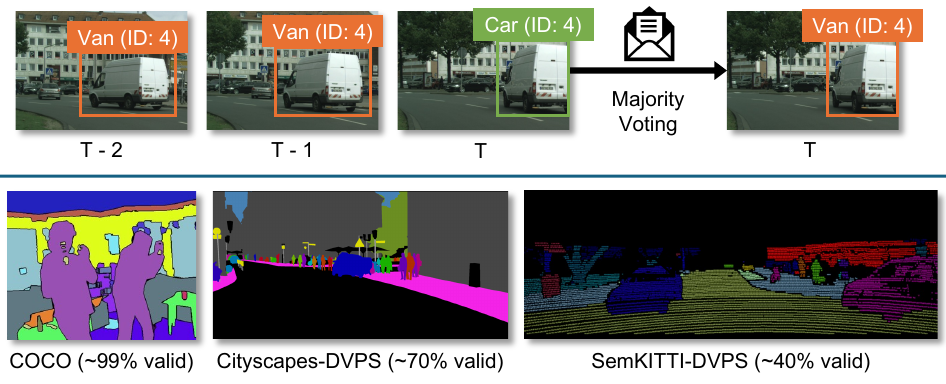}
    \vspace{-15pt}
    \caption{
        \textbf{Online majority voting (OMV) and point sampling.}
        The upper part illustrates our proposed OMV, which conducts class-agnostic instance association and refines the current frame classification based on the majority of the trajectory classes.
        The lower part shows the annotation difference among the COCO~\cite{kirillov2019panoptic} and DVPS datasets~\cite{qiao2021vip}, where black color stands for no ground truth.
    }
    \vspace{-10pt}
    \label{fig:rps_omv}
\end{figure}

\subsection{Online Tracking and Majority Voting}
\label{sec:method:tracking}
To track the instances, \ie thing masks, over time in an online manner, we associate the active track $\uptau_{t-1}$ and the current frame thing masks $\mathbf{M}^\text{thing}_{t}$.
We solve the association problem by using a weighted bipartite matching algorithm.
Unlike Uni-DVPS~\cite{jiyeon2024unidvps}, which directly uses instance queries as appearance features for matching, we design a lightweight tracking head with similarity learning (SL) to better distinguish instances.
Our tracking head is composed of one MLP block ($\mathrm{MLP}_{track}$) to project the final queries $\mathbf{q}_{l-1}$ and obtain the tracking queries $\mathbf{q}_{t}$ as instance embeddings for frame $t$ as:
\begin{equation}
    \mathbf{q}_{t} = \mathrm{MLP}_{track}(\mathbf{q}_{l-1}).
\end{equation}
We build the bi-directional softmax similarity matrix $\mathbf{A}$ between $\mathbf{q}_{t}$ of the current frame and the tracking queries $\mathbf{q}_{\uptau}$ of $\uptau_{t-1}$.
In particular, $\mathbf{A}$ is defined as:
\begin{equation}
    \mathbf{A} = \frac{1}{2} \left[ \frac{\exp(\mathbf{q}_{t} \cdot \mathbf{q}_{\uptau})}{\sum_{\mathbf{q}_{t}} \exp (\mathbf{q}_{t} \cdot \mathbf{q}_{\uptau})} + \frac{\exp (\mathbf{q}_{t} \cdot \mathbf{q}_{\uptau})}{\sum_{\mathbf{q}_{\uptau}} \exp(\mathbf{q}_{t} \cdot \mathbf{q}_{\uptau})} \right].
\end{equation}

If the instance query is matched to one of the active tracks, \ie similarity score is greater than $0.3$, we assign the tracking ID to the current instance query; otherwise, we create a new track for the instance.
After instance association, we update the active track as $\uptau_{t}$ and obtain the instance id for each mask for VPS.
In contrast to PolyphonicFormer~\cite{yuan2022polyphonicformer}, \ourmodel does not extract appearance embeddings based on RoI Align~\cite{he2017mask} on post-processed bounding boxes from instance masks.
Compared to Uni-DVPS~\cite{jiyeon2024unidvps}, our design enables online association with comparable performance or better results for long input sequences.

Moreover, to leverage the tracking ability of \ourmodel, we propose an online majority voting (OMV) mechanism to leverage the tracking ability and propagate temporal information across time, thereby refining class predictions as shown in \cref{fig:rps_omv}.
Unlike previous works~\cite{pang2021quasi,hu2022monocular,fischer2022cc,fischer2023qdtrack} that only associate within the same-class instances, we follow TETer~\cite{li2022tracking} and conduct class-agonistic instance association.
In contrast to TETer, which corrects the class after tracking through the entire sequence, we refine the current tracking frame based on the majority voting results over the classifications from the past five frames, allowing \ourmodel to be online.

\subsection{Training Losses}
\label{sec:method:loss}
\noindent{}\textbf{Panoptic Segmentation.}
We exploit a bipartite matching algorithm to align the predicted mask and the ground truth (GT).
After obtaining the prediction and GT pairs, we compute the cross-entropy loss ($\mathcal{L}_{ce}$) and the dice loss~\cite{milletari2016v} ($\mathcal{L}_{dice}$) for mask predictions.
For per-mask classification, we utilize cross-entropy loss ($\mathcal{L}_{cls}$) as the supervision.
The final loss for panoptic segmentation prediction $\mathcal{L}_{s} = \mathcal{L}_{s} = \lambda_{ce}\mathcal{L}_{ce} + \lambda_{dice}\mathcal{L}_{dice} + \lambda_{cls}\mathcal{L}_{cls}$,
where $\lambda_{ce}$ and $ \lambda_{dice}$ are both set to $5.0$ and $ \lambda_{cls}$ is $2.0$ as~\cite{cheng2022masked,jiyeon2024unidvps}.
We apply $\mathcal{L}_{s}$ to the segmentation output of each transformer decoder layer and initial queries to help gradually refine the features.

\noindent{}\textbf{Point Sampling.}
Mask2Former~\cite{cheng2022masked} proposes randomly sampling points from the mask prediction and ground truth for mask loss (point sampling), rather than using the entire mask (mask-based), to save three times the computational resources for training.
However, point sampling is only feasible for the fine-grained annotated datasets.
As shown in \cref{fig:rps_omv}, the autonomous driving datasets are usually noisier and contain many more invalid points.
The common practice is to use the mask-based approach while removing the unannotated pixels, but it is not feasible to fit the model on a $24$ GB GPU using mask-based mask loss for the high-resolution datasets like Cityscapes-DVPS.

To still benefit from the efficiency of the point sampling method, we remove invalid points from the point sampling results and noise from the training data.
We first follow \cite{cheng2022masked} to sample $K$ points for training, but filter out the invalid points during bipartite matching and computing losses.
This allows us to achieve the same performance as using full mask supervision with invalid pixels removed, while benefiting from the point sampling method and a fit batch size of $2$ on a single RTX 4090 with input resolutions of $1024\times2048$ for training.

\noindent{}\textbf{Depth Estimation.}
We use Scale-invariant loss~\cite{eigen2014depth} and absolute relative error $\mathcal{L}_{abs}$ to supervise the depth estimation.
The depth loss $\mathcal{L}_{depth}$ between ground truth $y^*$ and prediction $\hat{y}$ is formulated as
\begin{equation}
    \mathcal{L}_{depth} = \lambda_{depth}\sqrt{\mathbb{V}[\varepsilon] + \alpha\mathbb{E}^2[\varepsilon]} + \lambda_{abs}\mathcal{L}_{abs},
\end{equation}
where $\varepsilon = \log(y^*) - \log(\hat{y})$, and $\mathbb{V}[\varepsilon]$ and $\mathbb{E}[\varepsilon]$ are computed as the empirical variance and expected value over all valid pixels.
We set $\alpha, \lambda_{depth}, \lambda_{abs}$ to $0.15, 100.0, 10.0$, respectively.

\noindent{}\textbf{Instance Tracking.}
We use quasi-dense similarity learning~\cite{pang2021quasi, hu2022monocular, fischer2023qdtrack, fischer2022cc} to train our tracking head.
For every input key frame, the corresponding reference frame is sampled within a temporal interval $\Delta\mathrm{t} \sim \mathbf{U}[-3, 3]$.
The tracking queries presented in the key frame are noted as $\mathbf{q}_{t}^{key}$ and those in the reference frame as $\mathbf{q}_{t}^{ref}$.
We match the positive and negative samples across time through bipartite matching between the mask prediction and the ground truth.
Tracking queries are optimized by the multi-positive cross-entropy loss defined as follows:
\begin{equation}
     \mathcal{L}_{embed} = \log [1 + \sum_{\mathbf{q}_{p}^{ref}} \sum_{\mathbf{q}_{n}^{ref}} \text{exp}(\mathbf{q}_{t}^{key} \cdot \mathbf{q}_{n}^{ref} - \mathbf{q}_{t}^{key} \cdot \mathbf{q}_{p}^{ref})].
\end{equation}
The loss enforces each key frame tracking query to be similar to its positive reference $\mathbf{q}_{p}^{ref}$ and dissimilar to all its negative reference embeddings $\mathbf{q}_{n}^{ref}$.
We use the cosine similarity between the tracking queries in the key frame and the ones in the reference frame as an auxiliary loss:
\begin{equation}
    \mathcal{L}_{aux} = (\frac{\mathbf{q}_{t}^{key} \cdot \mathbf{q}_{t}^{ref}}{||\mathbf{q}_{t}^{key}|| \cdot ||\mathbf{q}_{t}^{ref}||} - \varepsilon)^2,
\end{equation}
where $\varepsilon$ is $1$ if key frame and reference frame queries are matched to the same ground truth object and $0$ otherwise.
The overall tracking loss $\mathcal{L}_{track} = \mathcal{L}_{embed} + \lambda_{embed}\mathcal{L}_{aux}$, where $\lambda_{embed}$ is set to 0.25.

\section{Experiements}
\label{sec:exp}

We first describe the details of datasets and evaluation metrics in \cref{sec:result:datasets} and \cref{sec:result:metrics}.
Then, we describe the implementation details in \cref{sec:result:implementation}, show the benchmark results in \cref{sec:result:benchmark} and analyze the results of ablation studies in \cref{sec:result:ablation}.
Finally, we show the qualitative comparison in \cref{sec:result:qualitative_comparison} and qualitative results of \ourmodel in \cref{sec:result:qualitative_results}.

\begin{table*}[t]
    \small
    \footnotesize
    \centering
    \caption{
        \textbf{Comparison with state-of-the-art methods.}
        Each cell shows DVPQ | DVPQ-Thing | DVPQ-Stuff under different window sizes.
        \ourmodel outperforms all the existing methods using RestNet-50 as the backbone.
    }
    \vspace{-5pt}
    \resizebox{1\linewidth}{!}{
    \begin{tabular}{l|ccc|ccc|ccc|ccc|ccc}
        \toprule
        \textbf{Cityscapes-DVPS} & \multicolumn{3}{c|}{k = 1} & \multicolumn{3}{c|}{k = 2} & \multicolumn{3}{c|}{k = 3} & \multicolumn{3}{c|}{k = 4} & \multicolumn{3}{c}{Average} \\
        \midrule
        ViP-DeepLab~\cite{qiao2021vip} & 47.4 & 38.8 & 53.7 & 44.0 & 28.1 & 51.6 & 39.0 & 23.3 & 50.5 & 37.5 & 20.2 & 50.0 & 42.0 & 27.6 & 51.5 \\
        PolyphonicFormer~\cite{yuan2022polyphonicformer} & 54.4 & 47.0 & 59.8 & 48.1 & 35.9 & 57.0 & 45.5 & 30.9 & 56.2 & 44.1 & 28.6 & 55.4 & 48.1 & 35.6 & 57.1 \\
        MonoDVPS~\cite{petrovai2023monodvps} & 57.2 & 48.4 & 63.6 & 51.0 & 37.0 & 61.0 & 47.9 & 31.0 & 60.0 & 45.7 & 27.0 & 59.3 & 50.4 & 35.9 & 61.0 \\
        Uni-DVPS~\cite{jiyeon2024unidvps} & 58.0 & 48.1 & 65.2 & 52.4 & 38.3 & 62.7 & 49.5 & 32.9 & 61.5 & 47.4 & 29.2 & 60.7 & 51.8 & 37.1 & 62.5 \\
        Multiformer~\cite{stolle2025balancing} & - & - & - & - & - & - & - & - & - &- & - & - & 54.8 & 37.4 & \textbf{67.4} \\
        \midrule
        \textbf{\ourmodel (Ours)} & \textbf{63.0} & \textbf{55.8} & \textbf{68.2} & \textbf{56.4} & \textbf{44.1} & \textbf{65.3} & \textbf{52.9} & \textbf{38.4} & \textbf{63.5} & \textbf{50.4} & \textbf{33.8} & \textbf{62.4} & \textbf{55.7} & \textbf{43.0} & 64.9 \\
        \bottomrule
        \toprule
        \textbf{SemKITTI-DVPS} & \multicolumn{3}{c|}{k = 1} & \multicolumn{3}{c|}{k = 5} & \multicolumn{3}{c|}{k = 10} & \multicolumn{3}{c|}{k = 20} & \multicolumn{3}{c}{Average} \\
        \midrule
        MonoDVPS~\cite{petrovai2023monodvps} & 43.3 & 37.5 & 47.6 & 38.1 & 27.4 & 45.9 & 36.9 & 25.3 & 45.4 & 35.9 & 23.6 & 45.0 & 38.6 & 28.4 & 46.0 \\
        PolyphonicFormer~\cite{yuan2022polyphonicformer} & 44.8 & 39.9 & 48.3 & 40.0 & 31.1 & 46.5 & 38.7 & 28.8 & 45.8 & 37.8 & 27.7 & 45.2 & 40.3 & 31.9 & 46.5 \\
        Uni-DVPS~\cite{jiyeon2024unidvps} & 47.0 & 41.4 & 51.1 & 43.5 & 35.8 & 49.1 & 41.1 & 31.6 & 47.9 & 37.4 & 25.2 & 46.3 & 42.2 & 33.5 & 48.6 \\
        \midrule
        \textbf{\ourmodel (Ours)} & \textbf{49.6} & \textbf{44.9} & \textbf{53.0} & \textbf{46.3} & \textbf{40.3} & \textbf{50.7} & \textbf{45.1} & \textbf{38.6} & \textbf{49.8} & \textbf{44.1} & \textbf{37.5} & \textbf{48.9} & \textbf{46.3} & \textbf{40.3} & \textbf{50.6} \\
        \bottomrule
    \end{tabular}}
    \vspace{-5pt}
    \label{tab:results:benchmark}
\end{table*}

\subsection{Datasets}
\label{sec:result:datasets}
\noindent{}\textbf{Cityscapes-DVPS.}
Cityscapes~\cite{cordts2016cityscapes} contains image-level panoptic annotations with $19$ semantic classes, including $8$ thing and $11$ stuff classes.
Cityscapes-VPS~\cite{kim2020video} extends Cityscapes by annotating $5$ extra frames between each annotation and proposes a new video panoptic segmentation dataset.
ViP-DeepLab~\cite{qiao2021vip} further extends Cityscapes-VPS with disparity maps computed via stereo matching from the Cityscapes dataset and proposes the Cityscapes-DVPS dataset, which includes training, validation, and test sets with $2400$, $300$, and $300$ frames, respectively.

\noindent{}\textbf{SemKITTI-DVPS.}
SemanticKITTI~\cite{behley2019semantickitti} is based on the odometry split of the KITTI~\cite{geiger2012we} and provides both RGB images and synchronized point clouds annotated at panoptic-level with $8$ thing and $11$ stuff classes.
The dataset is split into $11$ training and $11$ test sequences, and the training sequence 08 is used as the validation set.
ViP-DeepLab projects the 3D point clouds into the image plane with the proposed disparity consistency check and builds the SemKITTI-DVPS dataset, which includes training, validation, and test sets with $19130$, $4071$, and $4342$ frames, respectively.

\subsection{Evaluation Metrics.}
\label{sec:result:metrics}
The evaluation metric of DVPS is Depth-aware video panoptic quality (DVPQ)~\cite{qiao2021vip}, which aims to evaluate panoptic quality (PQ)~\cite{kirillov2019panoptic} over time, \ie video panoptic quality (VPQ)~\cite{kim2020video}, and also considers depth prediction as the metric to compute the threshold for inlier samples.
More specifically, let $\mathbf{P}_i^c$, $\mathbf{P}_i^{id}$, and $\mathbf{P}_i^d$ denote the predictions of example $i$ on the semantic class, instance ID, and depth.
Similarly, $\mathbf{T}_i^c$, $\mathbf{T}_i^{id}$, and $\mathbf{T}_i^d$ stand for ground truth notation.
Let $k$ be the window size of the time sequence and $\lambda$ be the depth threshold.
Then, $\mathrm{DVPQ}_{\lambda}^k(\mathbf{P}, \mathbf{T})$ is defined as
\begin{equation}
    \mathrm{PQ}(
        \begin{bmatrix}
            \|^{t+k-1}_{i=t}({\mathbf{\hat{P}}_i^c, \mathbf{P}_i^{id}}), \|^{t+k-1}_{i=t}({\mathbf{T}_i^c, \mathbf{T}_i^{id}})
        \end{bmatrix}^{T-k+1}_{t=1}
    ),
\end{equation}
where $\mathbf{\hat{P}}_i^c = \mathbf{P}_i^c$ for pixels that have absolute relative depth errors under $\lambda$ (\ie $| \mathbf{P}_i^d - \mathbf{T}_i^d | / \mathbf{T}_i^d \leq \lambda$), and will be assigned a void label otherwise.

There are four different $k$'s, \eg $ k \in \{1, 2, 3, 4\}$ and $ k \in \{1, 5, 10, 20\}$ for Cityscapes-DVPS and SemKitti-DVPS, respectively.
Both datasets use the three values of $\lambda$, \eg $\lambda \in \{0.1, 0.25, 0.5\}$, which approximately correspond to the percentage of depth inlier metric $\delta$ < $1.1$, $\delta$ < $1.25$ and $\delta$ < $1.5$.
The final DVPQ number is obtained by averaging all the values of $k$ and $\lambda$.
The depth-aware panoptic quality (DPQ) metric can be directly obtained from the DVPQ calculation by setting $k = 1$ and averaging on all $\lambda$, while VPQ can be obtained by setting $\lambda$ to 0.
The DVPQ metric can be divided into thing classes (DVPQ-th) and stuff classes (DVPQ-st).

\subsection{Implementation Details}
\label{sec:result:implementation}
\ourmodel is implemented in PyTorch~\cite{paszke2019pytorch} and CUDA~\cite{nickolls2008cuda}.
We follow previous work~\cite{jiyeon2024unidvps} and use the same pre-trained ResNet-50~\cite{he2016deep} as the backbone and the same training schedule for a fair comparison.
We follow~\cite{qiao2021vip,yuan2022polyphonicformer,jiyeon2024unidvps} to train the PS model with only $\mathcal{L}_{s}$.
Then we further optimize \ourmodel for DPS on Cityscapes-DVPS and SemKITTI-DVPS using $\mathcal{L}_{s}$ and $\mathcal{L}_{depth}$, respectively.
Finally, we freeze the DPS model and train only the tracking head for similarity learning with $\mathcal{L}_{track}$ for $12$ epochs.
We use random color jittering, horizontal flipping, and large-scale jittering for PS training, color jittering, random rescaling, and horizontal flipping for DPS training, and random flipping for tracking training.
All experiments are conducted with 8 RTX 4090s and a batch size of $16$, using the AdamW optimizer~\cite{kingma2014adam} with an initial learning rate of $0.0001$.

\subsection{Comparison with State-of-the-art}
\label{sec:result:benchmark}
\noindent{}\textbf{Cityscapes-DVPS.}
\cref{tab:results:benchmark} shows that \ourmodel achieves a new state-of-the-art (SOTA) performance on the Cityscapes-DVPS benchmark.
Our design for the DPS module outperforms Uni-DVPS by $5$ points when $k=1$, while our online tracking mechanism still outperforms their offline method by $3$ points when $k=4$.
This results in a general $\textbf{3.9}$ point increase over the previous open-sourced state-of-the-art (SOTA) and $\textbf{0.7}$ over MultiFormer.
As shown in \cref{tab:results:fps}, \ourmodel can also achieve higher FPS at the same time, thus lying on the Pareto optimal frontier of efficiency \vs performance.

\noindent{}\textbf{SemKITTI-DVPS.}
SemKITTI-DVPS is more challenging than Cityscapes-DVPS due to its sparser ground-truth annotation from the projected LiDAR point cloud and the longer validation sequence.
As shown in \cref{tab:results:benchmark}, our design consistently outperforms existing methods and achieves a new SOTA.
It is worth noting that our online mechanism is more suitable for long tracking sequences.
As shown in \cref{tab:results:fps}, Uni-DVPS needs to run different sequence lengths to obtain the final results, and the longer the sequence, the lower the FPS.
In contrast, \ourmodel only needs one pass to get the final results on SemKITTI-DVPS with high FPS.
Moreover, when $k=1$, we outperform Uni-DVPS by $2.6$ points, while when $k=20$, we outperform them by $7.3$ points.
We successfully associate instances over time and obtain noticeably higher results in the DVPQ-Thing metric, which, in turn, leads to an overall improvement in the DVPQ.

\begin{table}[t]
    \small
    \footnotesize
    \centering
    \caption{
        \textbf{FPS Comparison on RTX 4090.}
        ``-'' stands for the implementation is unavailable.
        Uni-DVPS requires running $k = 1, 5, 10, 20$ respectively for SemKITTI-DVPS, while the FPS decreases with more frames.
    }
    \vspace{-5pt}
    \resizebox{1\linewidth}{!}{
    \begin{tabular}{lcc}
        \toprule
        Method & Cityscapes-DVPS ($1024 \times 2048$) & SemKITTI-DVPS ($384 \times 1280$) \\ 
        \midrule
        PolyphonicFormer~\cite{yuan2022polyphonicformer} & 1.8 & - \\
        Uni-DVPS~\cite{jiyeon2024unidvps} & 9.2 & 44.7 / 10.2 / 5.1 / 2.6 \\
        \textbf{Ours} & \textbf{12.8} & \textbf{46.9} \\
        \bottomrule
    \end{tabular}
    }
    \label{tab:results:fps}
    \vspace{-5pt}
\end{table}
\begin{table}[t]
    \small
    \footnotesize
    \centering
    \caption{
        \textbf{Depth head comparison.}
        We show that our one-pass depth estimation head is faster for both training and inference.
    }
    \vspace{-5pt}
    \resizebox{1\linewidth}{!}{
    \begin{tabular}{lcc}
        \toprule
        Method & Training steps per second $\uparrow$ & Depth inference time (ms) $\downarrow$  \\ 
        \midrule
        Uni-DVPS & 1.2 & 1.83 \\
        \textbf{ESD (Ours)} & \textbf{1.4} & \textbf{0.72} \\
        \bottomrule
    \end{tabular}
    }
    \vspace{-5pt}
    \label{tab:results:depth_comp}
\end{table}

\subsection{Ablation Study}
\label{sec:result:ablation}
We conduct the ablation studies on Cityscapes-DVPS to validate each of our contributions.
We use ResNet-50 as the backbone and use the same training setting across all experiments.
We first analyze our designs for the DPS module in \cref{tab:results:ablation_dps} by gradually adding our proposed design to the baseline (row 1), \ie Uni-DVPS.

\noindent{}\textbf{Explicit Scene Discretization.}
As shown in \cref{tab:results:depth_comp} and row 2 of \cref{tab:results:ablation_dps}, by only replacing the per-segment depth prediction with our proposed explicit scene discretization (ESD), we can have faster both training and inference speed while successfully improve the PQ and $\delta < 1.1$, resulting in $1.7$ higher DPQ.
This suggests that our ESD can better leverage explicit scene representation to obtain improved depth, and also help segmentation with depth supervision.
We further remove the ESD from the full method (row 5), which leads to worse depth estimation results and segmentation quality, supporting its effectiveness.

\noindent{}\textbf{Separate Pixel Representation.}
Due to the difference in the output spaces for MMDE and PS, we further disentangle the pixel representation into $\mathrm{E}_s$ and $\mathrm{E}_d$ (row 3 of \cref{tab:results:ablation_dps}).
The results show that after separating the representation, it will consistently improve the panoptic quality and depth estimation, leading to an overall improvement of $1.0$ DPQ.

\noindent{}\textbf{Point Sampling.}
In row 4 of \cref{tab:results:ablation_dps}, the results show that addressing the annotation quality issue in existing DVPS datasets can further improve panoptic quality and produce an additional improvement of $1.8$ in DPQ.

We analyze our designs for tracking in \cref{tab:results:ablation_track}, where all experiments are online.
We follow PolyphonicFormer~\cite{yuan2022polyphonicformer} and build the DVPS baseline (row 1) for our final DPS model, which converts masks as 2D bounding boxes and extracts appearance features with an extra RoI feature extractor.

\begin{table}[t]
    \footnotesize
    \small
    \centering
    \caption{
        \textbf{Ablations of depth and segmentation.}
        Sep. Pixel stands for separate pixel representation.
        We report the inlier metric $\delta < 1.1$ for depth.
        PS$^*$ denotes remove invalid pixels during point sampling.
    }
    \vspace{-5pt}
    \resizebox{1\linewidth}{!}{
    \begin{tabular}{lccc|ccc|c|c}
        \toprule
        & ESD & Sep. Pixel & PS$^*$ & DPQ $\uparrow$ & DPQ-th $\uparrow$ & DPQ-st $\uparrow$ & PQ $\uparrow$ & $\delta < 1.1$ (\%) $\uparrow$ \\ 
        \midrule
        1 & - & - & - & 58.0 & 48.1 & 65.2 & 65.9 & 78.3 \\
        2 & \checkmark & - & - & 59.7 & 50.5 & 66.4 & 67.7 & 79.3 \\
        3 & \checkmark & \checkmark & - & 60.7 & 52.9 & 66.3 & 68.0 & 79.5 \\
        4 & \checkmark & \checkmark & \checkmark & \textbf{62.5} & \textbf{54.5} & \textbf{68.4} & \textbf{69.7} & \textbf{79.5} \\
        \midrule
        5 & - & \checkmark & \checkmark & 60.2 & 51.0 & 66.8 & 68.0 & 78.2 \\
        \bottomrule
    \end{tabular}
    }
    \vspace{-5pt}
    \label{tab:results:ablation_dps}
\end{table}
\begin{table}[t]
    \footnotesize
    \small
    \centering
    \caption{
        \textbf{Tracking ablations study of \ourmodel.}
        Query stands for using instance queries for tracking, and SL stands for similarity training.
    }
    \vspace{-5pt}
    \resizebox{1\linewidth}{!}{
    \begin{tabular}{lccc|cc|cc|cc}
        \toprule
         & Query & SL & OMV & DVPQ & DVPQ-th $\uparrow$ & VPQ $\uparrow$ & VPQ-th $\uparrow$ & PQ $\uparrow$ & PQ-th $\uparrow$ \\ 
        \midrule
        1 & - & - & - & 54.3 & 39.7 & 60.6 & 44.8 & \multirow{3}{*}{69.7} & \multirow{3}{*}{61.4} \\
        2 & \checkmark & - & - & 54.7 & 40.7 & 61.2 & 46.0 & & \\
        3 & \checkmark & \checkmark & - & 55.2 & 41.8 & 61.8 & 47.5 & & \\
        \midrule
        4 & \checkmark & \checkmark & \checkmark & 55.7 & 43.0 & 62.2 & 48.7 & 70.1 & 62.7 \\
        \bottomrule
    \end{tabular}
    }
    \label{tab:results:ablation_track}
    \vspace{-5pt}
\end{table}

\noindent{}\textbf{Query Association.}
We validate the effectiveness of using transformer queries for tracking.
Without any similarity training (SL), directly using the transformer queries for online tracking (row 2) can already yield better results than an extra feature extractor.
We speculate that it is due to the imprecise conversion from masks to boxes, which not only causes extra computational effort but also generates a non-ideal appearance.
We further validate SL with a lightweight tracking head for transformer queries (row 3), which results in a $1.5$ and $0.5$ points improvement in VPQ-th and DVPQ, respectively.

\noindent{}\textbf{Online Majority Voting.}
In row 4, we further test the effectiveness of the proposed online majority voting (OMV).
Our improved association ability leads to a $0.4$  improvement in PQ and a gain of $1.2$ VPQ-th, leading to a $0.5$-point higher overall DVPQ.

\subsection{Qualitative Comparison}
\label{sec:result:qualitative_comparison}
\begin{figure}[t]
    \small
    \footnotesize
    \centering
    \setlength\tabcolsep{5pt}
    \begin{tabular}{ccc}
        \includegraphics[width=0.3\linewidth]{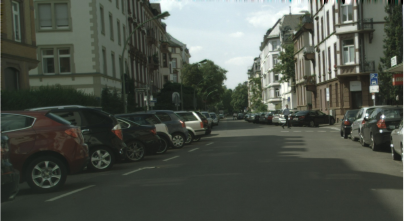} &
        \includegraphics[width=0.3\linewidth]{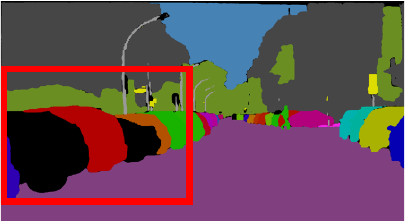} &
        \includegraphics[width=0.3\linewidth]{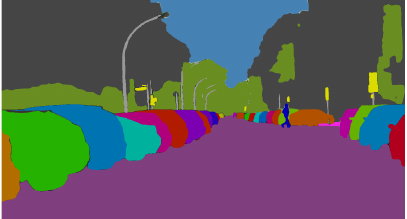} \\
        \includegraphics[width=0.3\linewidth]{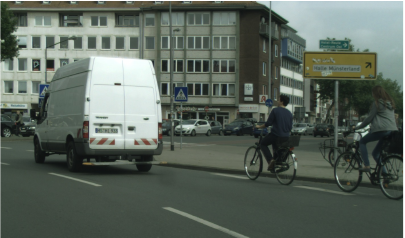} &
        \includegraphics[width=0.3\linewidth]{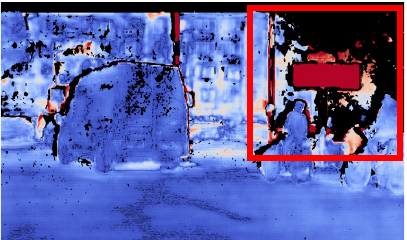} &
        \includegraphics[width=0.3\linewidth]{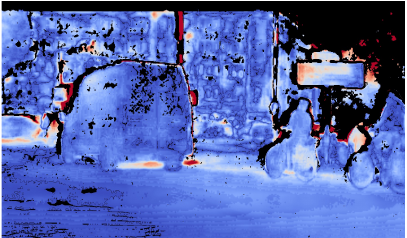} \\
        Input RGB & Uni-DVPS & Ours \\
    \end{tabular}
    \vspace{-5pt}
    \caption{
        \textbf{Qualitative analyses.}
        We compare Uni-DVPS with our method on Cityscapes-DVPS for both segmentation and absolute relative error for depth
        The red boxes highlight the inaccuracies of the previous SOTA.
    }
    \label{fig:sota_comparison}
    \vspace{-10pt}
\end{figure}

We directly compare the qualitative results between \ourmodel and the previous SOTA, \ie Uni-DVPS~\cite{jiyeon2024unidvps} in \cref{fig:sota_comparison}.
We achieve better instance tracking and geometric understanding, which leads to better DVPS results.

\begin{figure*}[t]
    \centering
    \small
    \footnotesize
    \scriptsize
    \resizebox{1\linewidth}{!}{
    \begin{tabular}{cccc} 
        \includegraphics[width=0.15\linewidth]{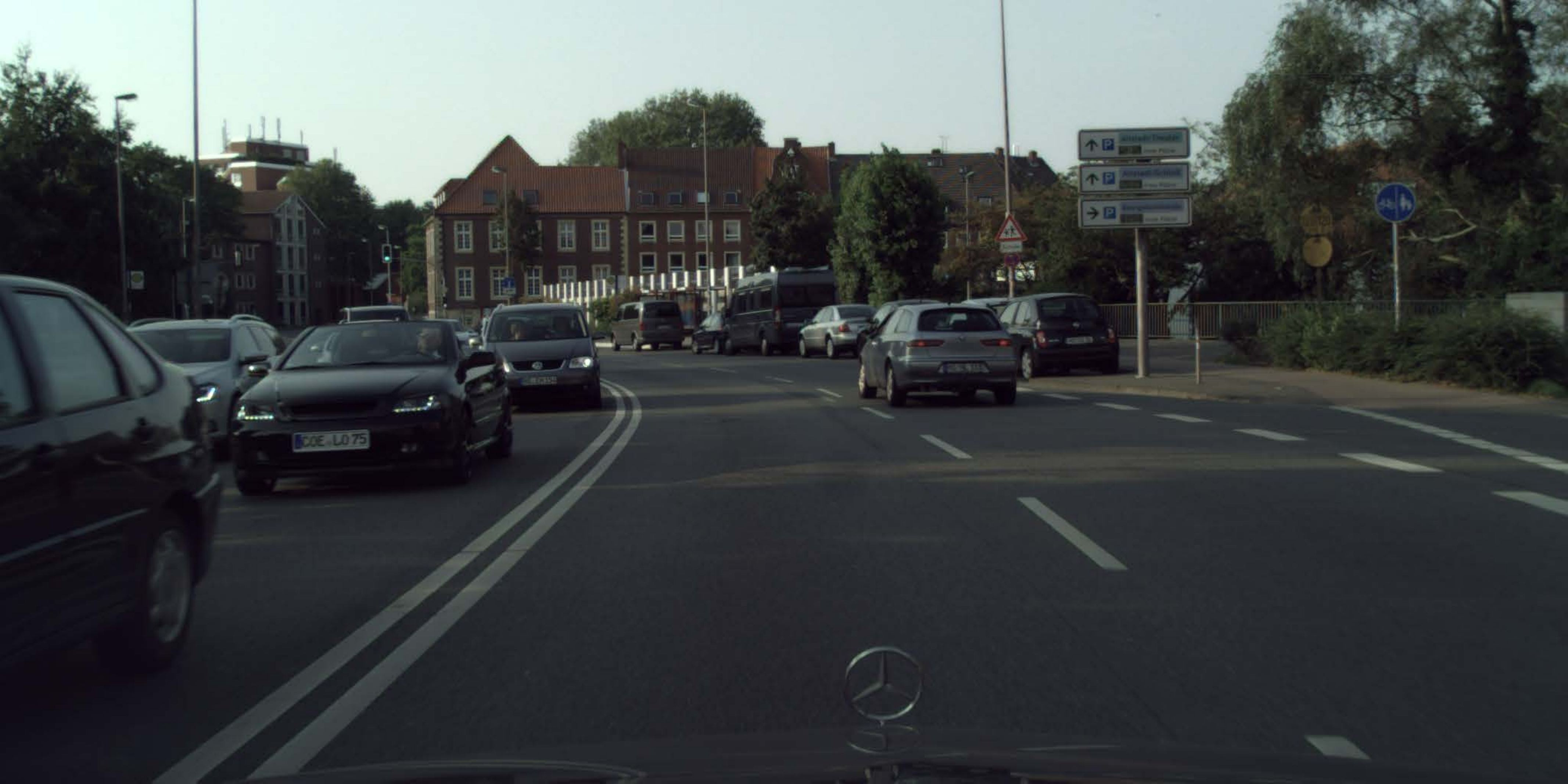}
        &
        \includegraphics[width=0.15\linewidth]{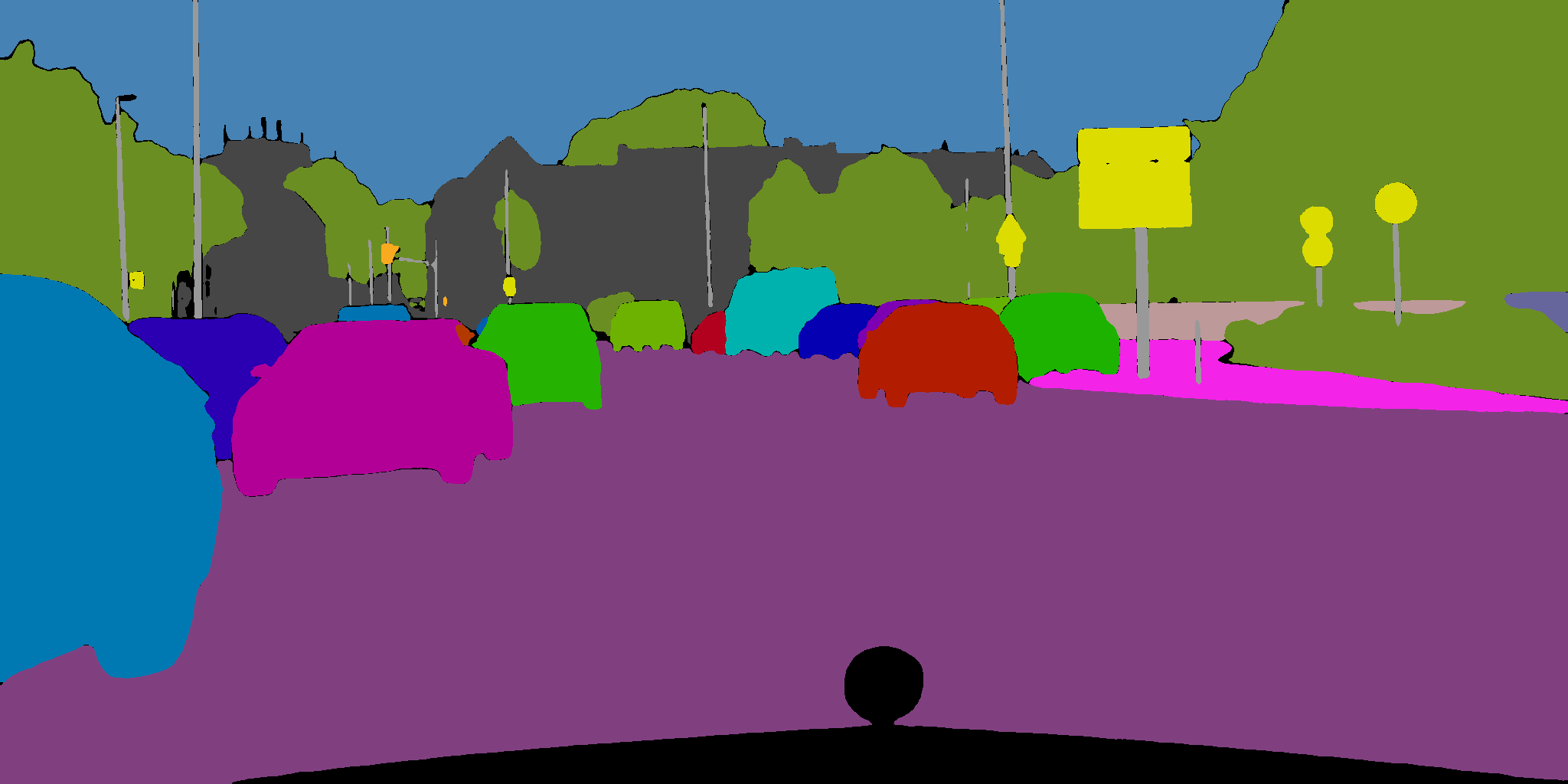}
        &
        \includegraphics[width=0.15\linewidth]{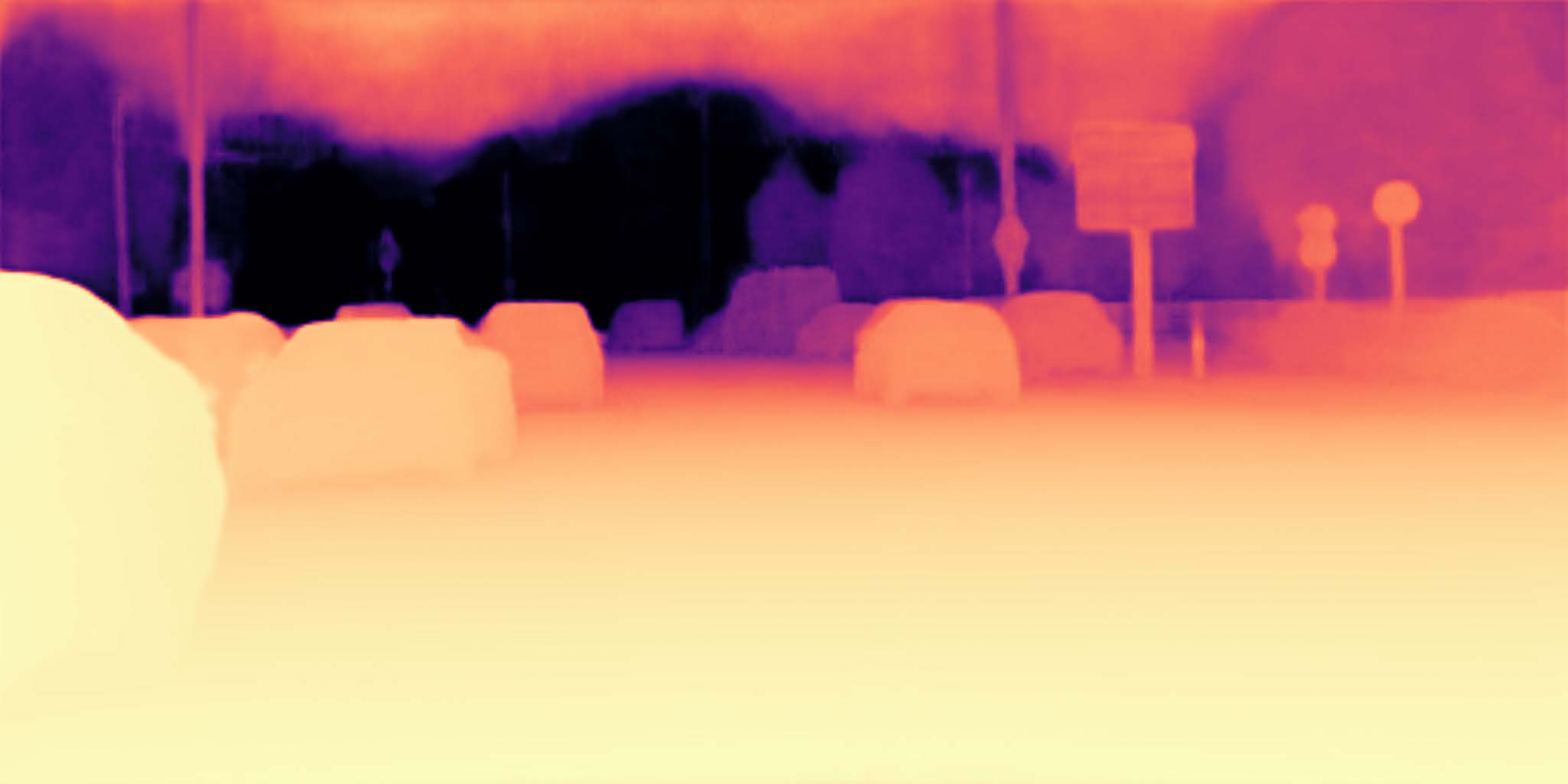}
        &
        \includegraphics[width=0.2\linewidth, height=1.7cm]{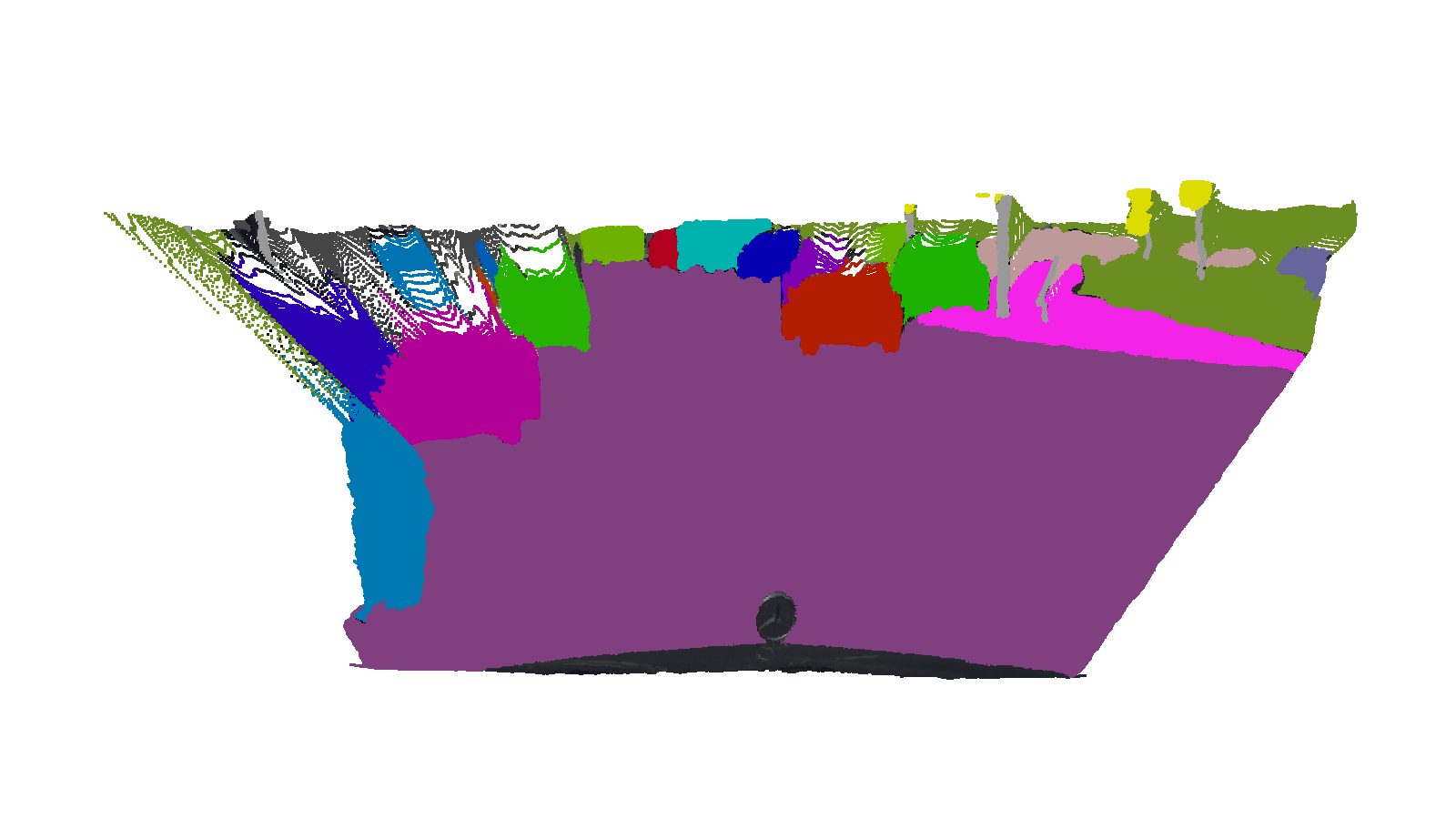}
        \vspace{-10pt}
        \\
        \includegraphics[width=0.15\linewidth]{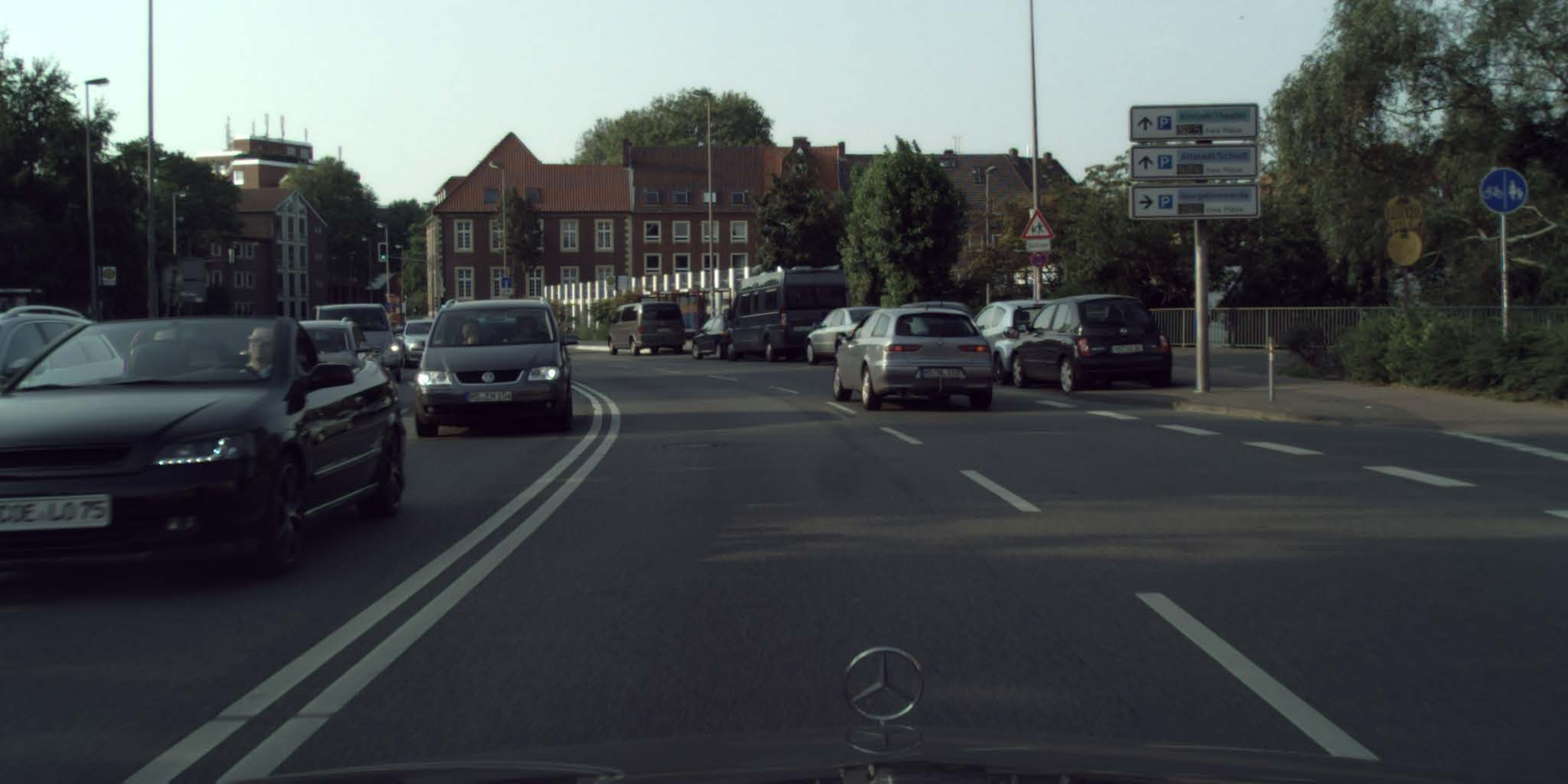}
        &
        \includegraphics[width=0.15\linewidth]{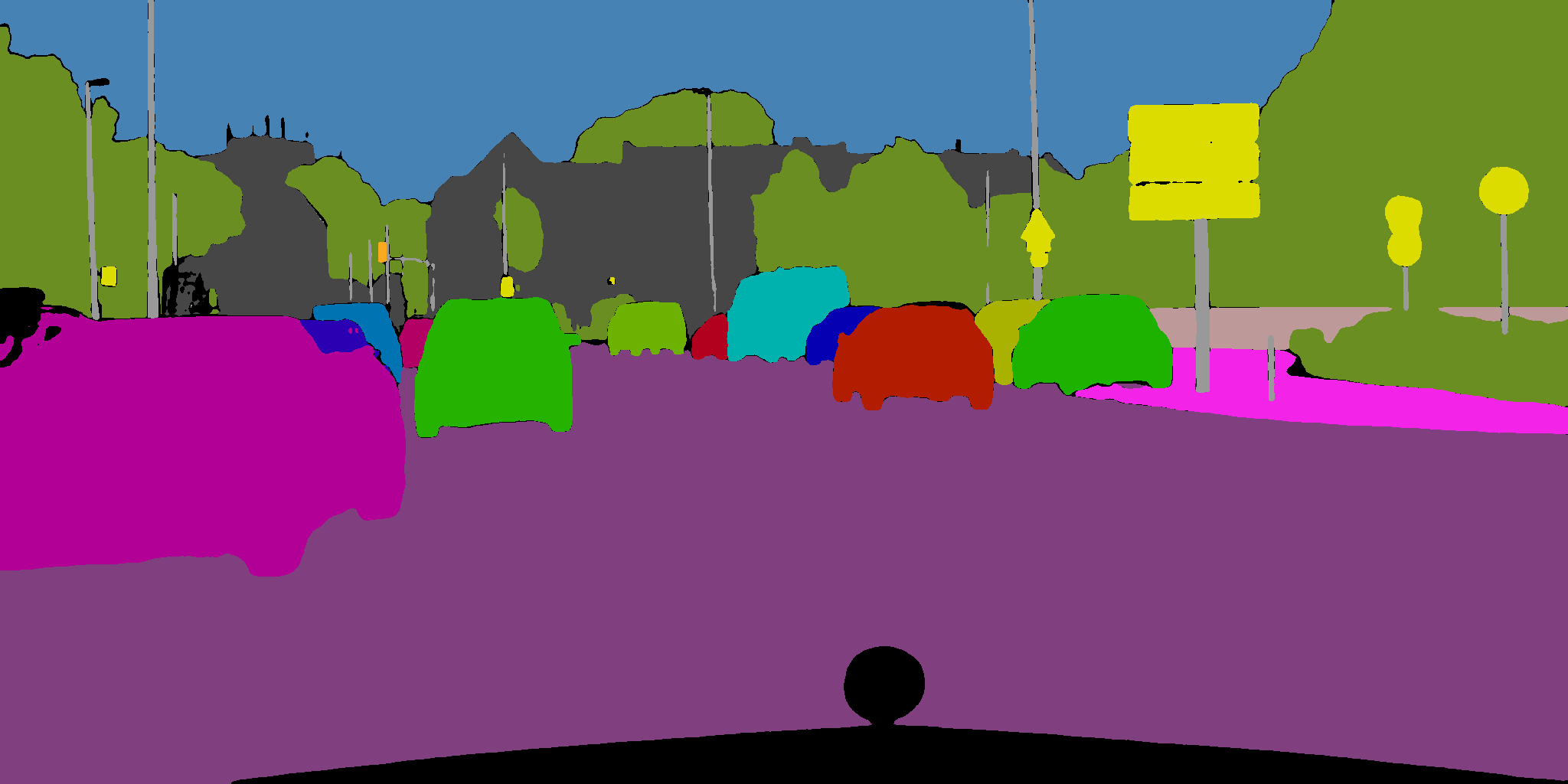}
        &
        \includegraphics[width=0.15\linewidth]{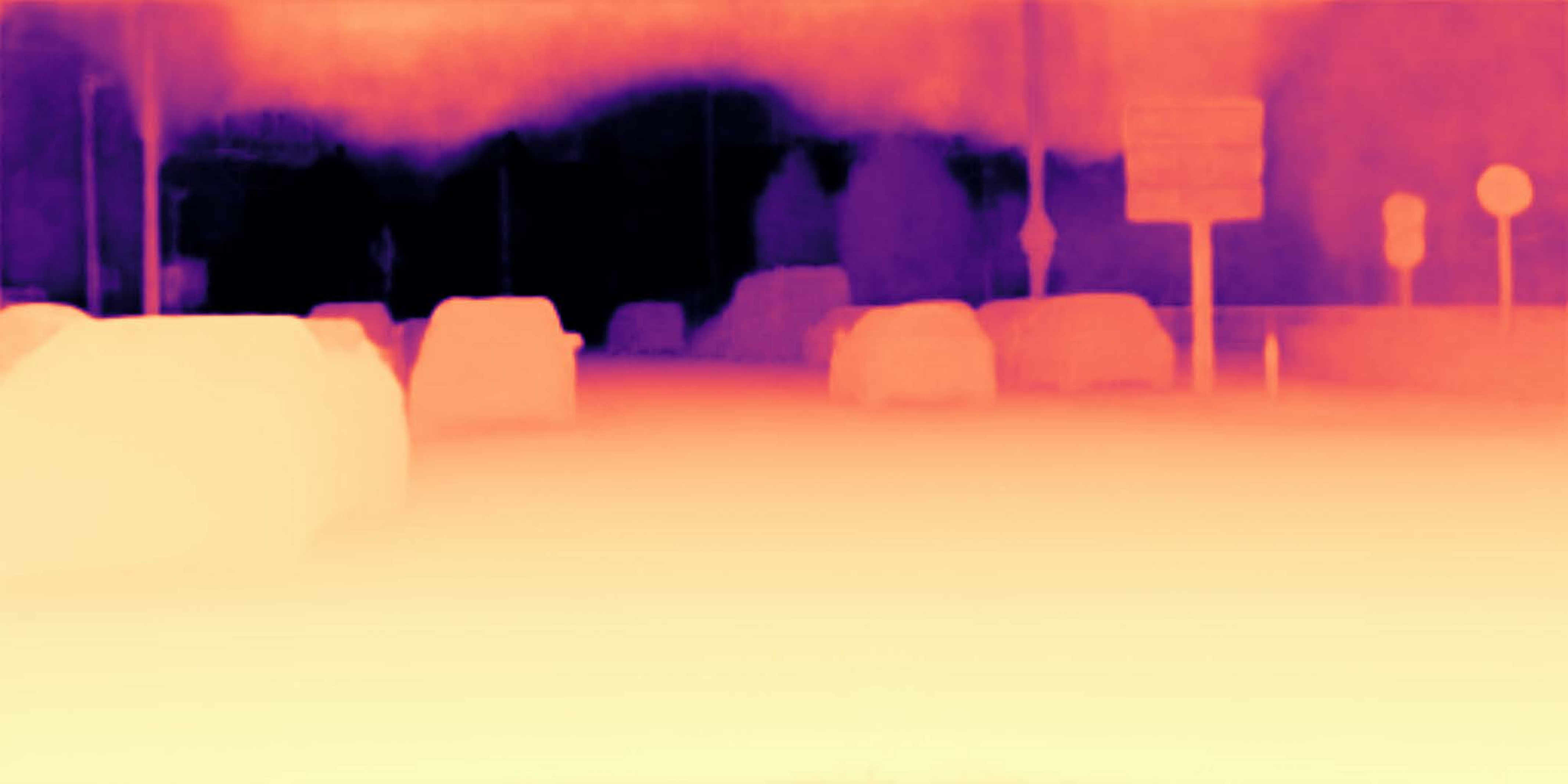}
        &
        \includegraphics[width=0.2\linewidth, height=1.7cm]{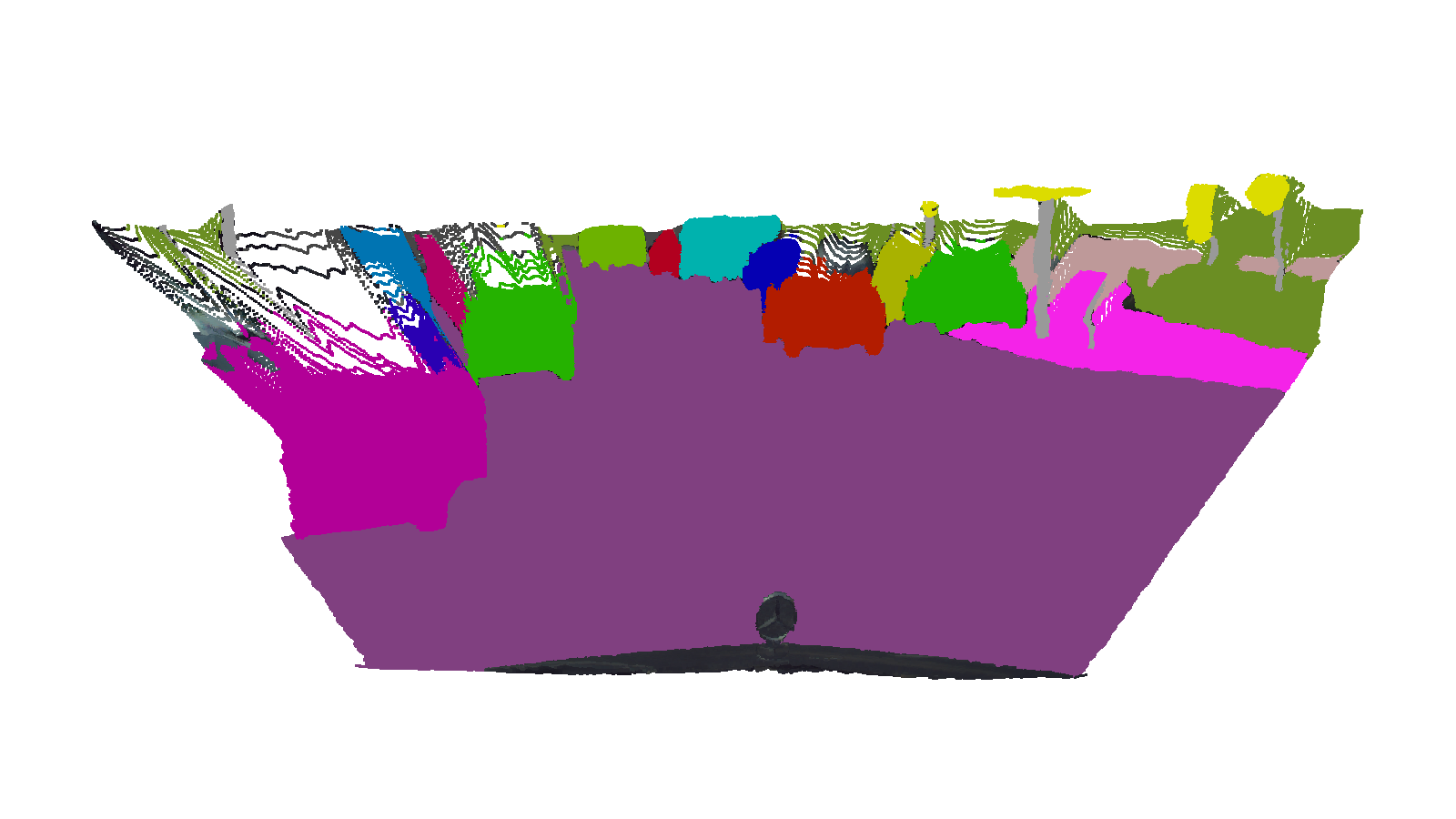}
        \vspace{-10pt}
        \\
        \includegraphics[width=0.15\linewidth]{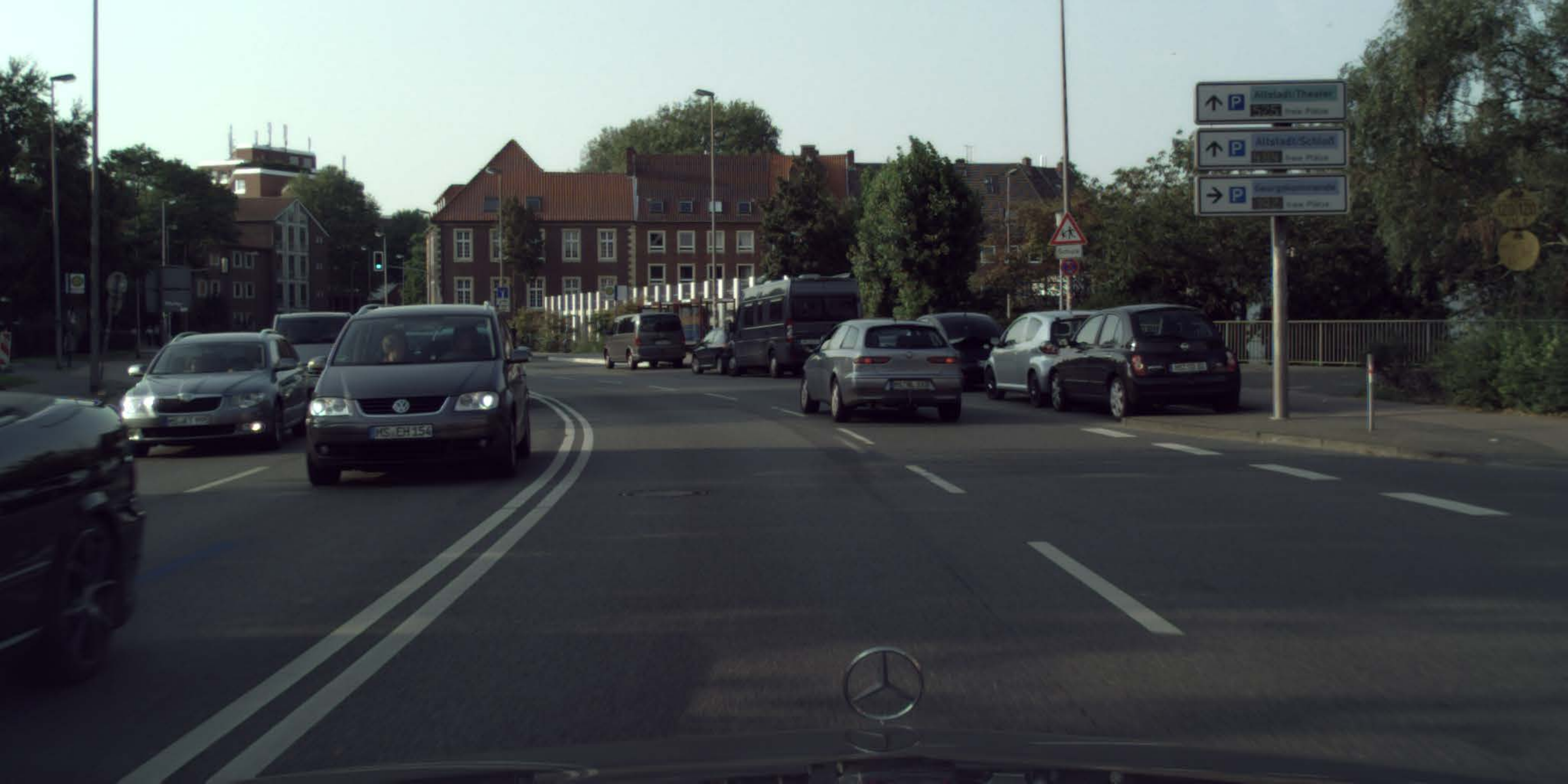}
        &
        \includegraphics[width=0.15\linewidth]{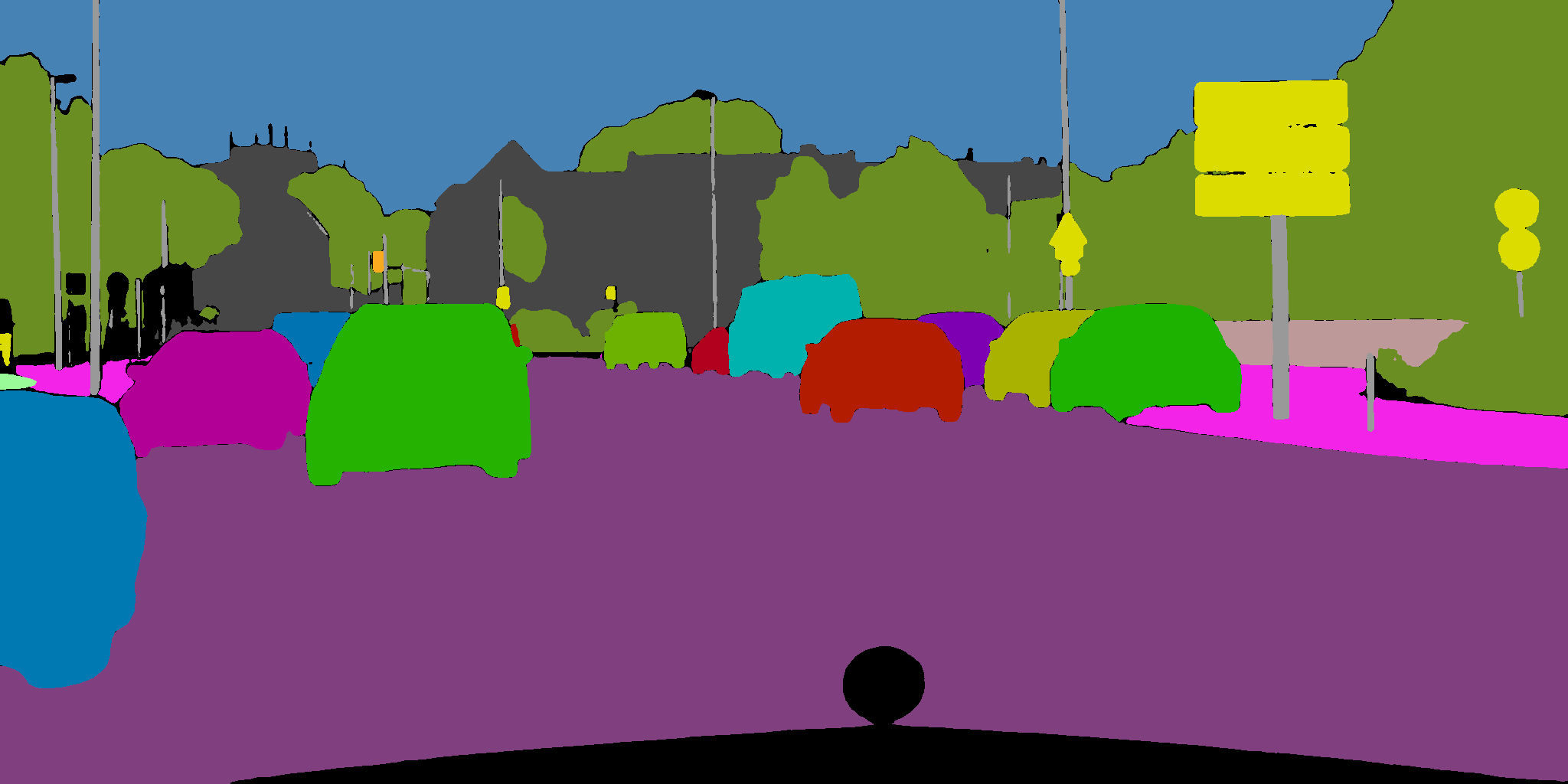}
        &
        \includegraphics[width=0.15\linewidth]{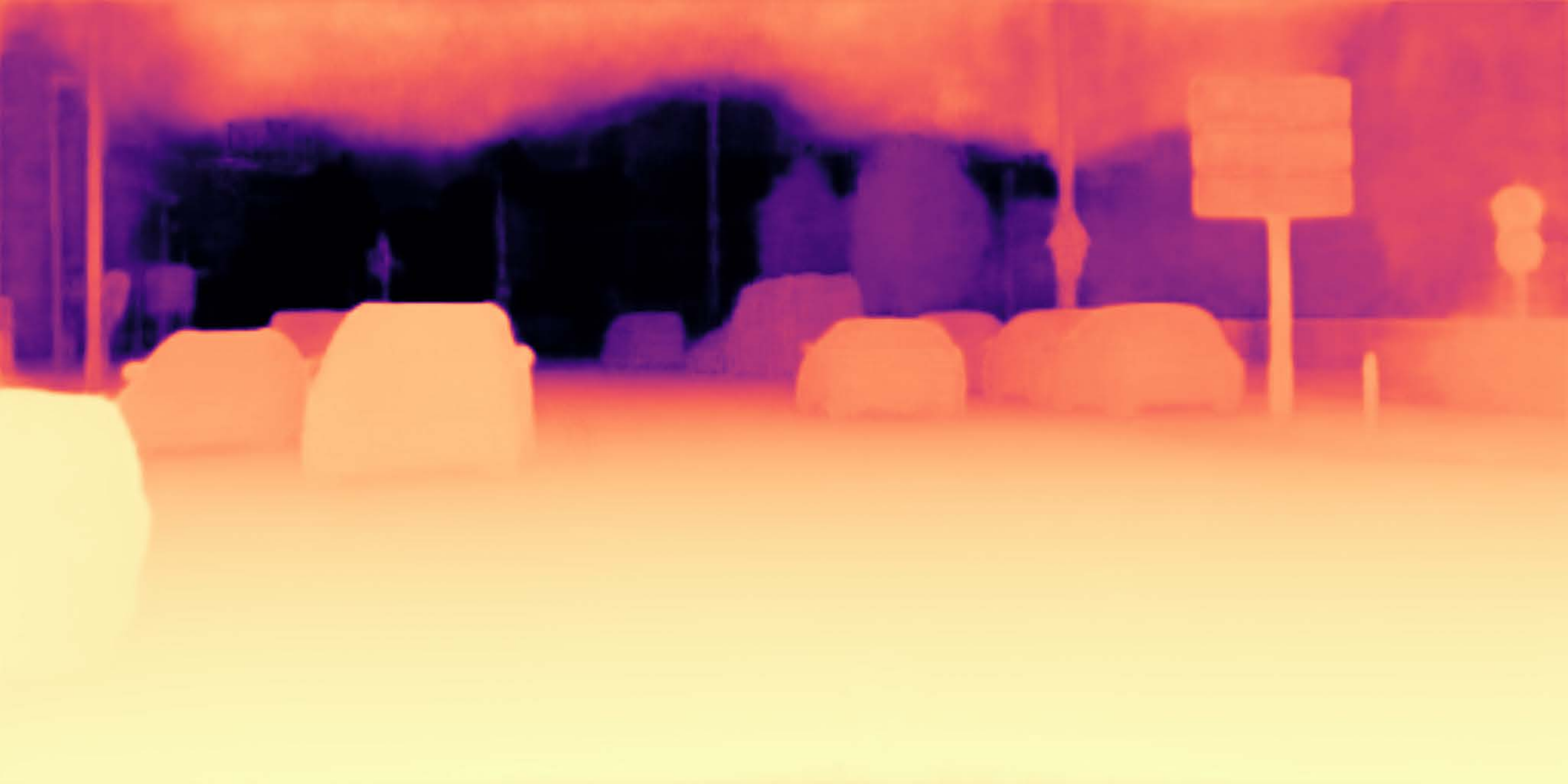}
        &
        \includegraphics[width=0.2\linewidth, height=1.7cm]{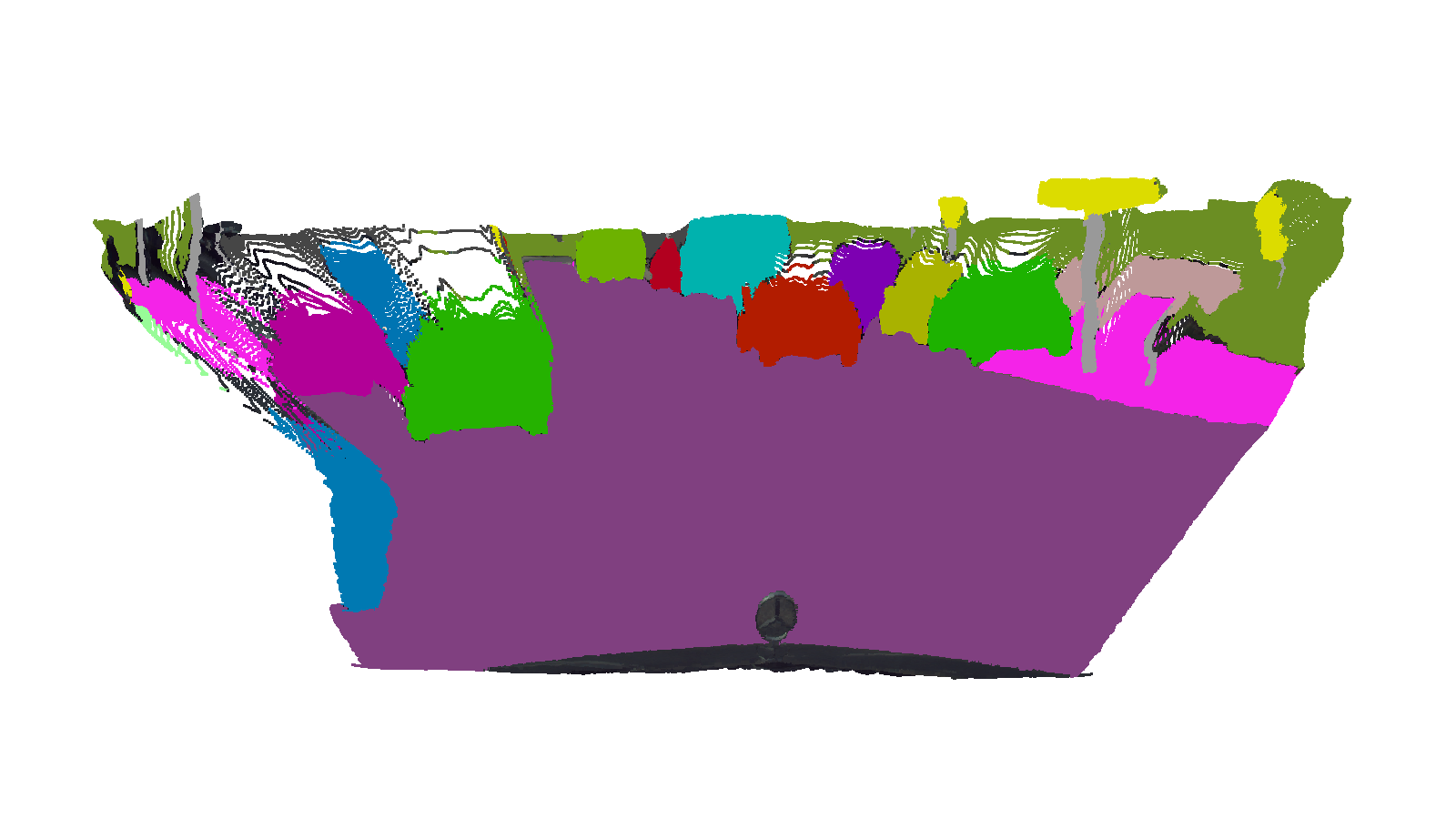}
        \\
        \midrule
        \includegraphics[width=0.15\linewidth]{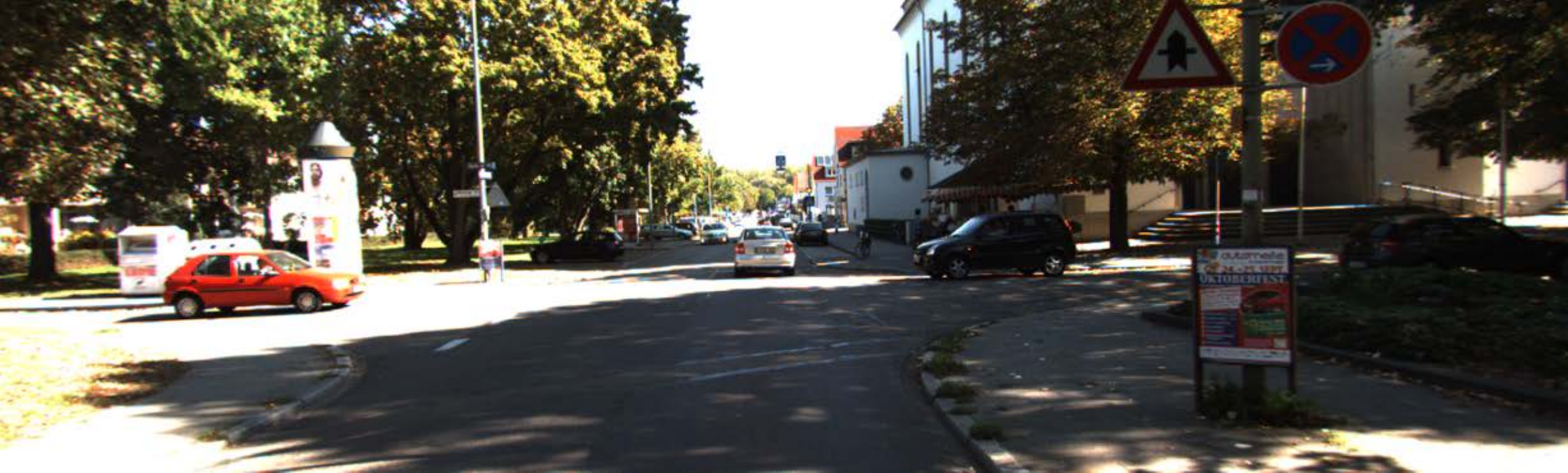}
        &
        \includegraphics[width=0.15\linewidth]{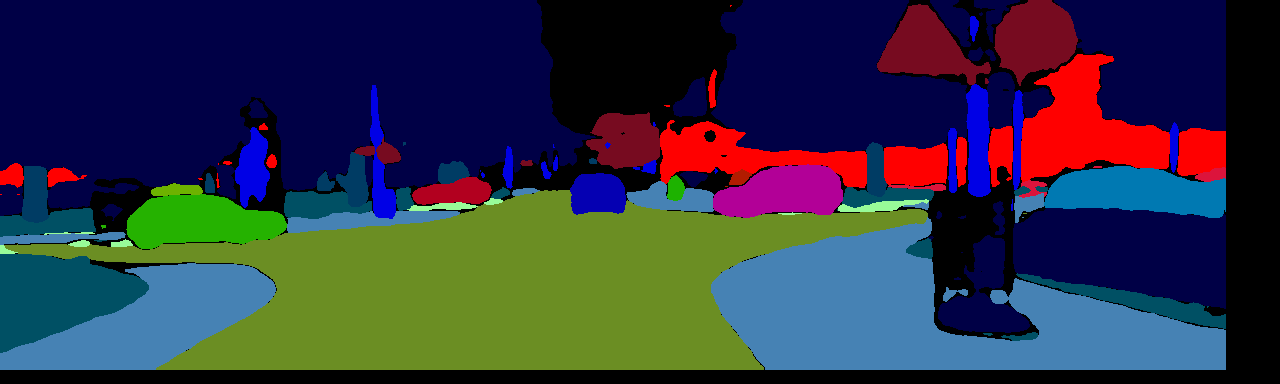}
        &
        \includegraphics[width=0.15\linewidth]{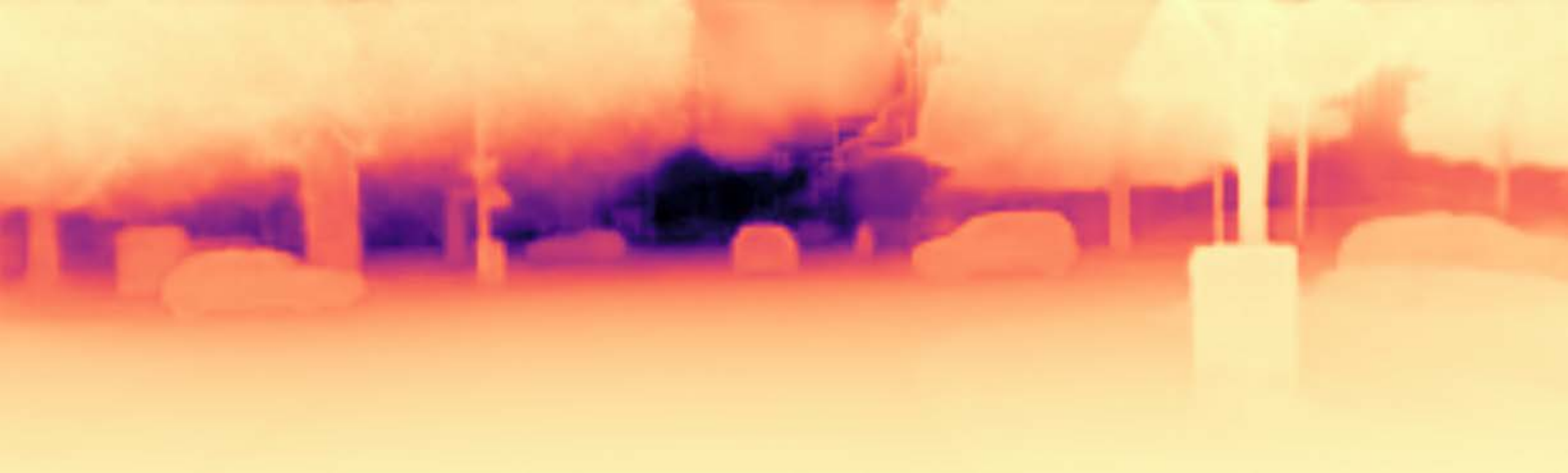}
        &
        \includegraphics[width=0.2\linewidth]{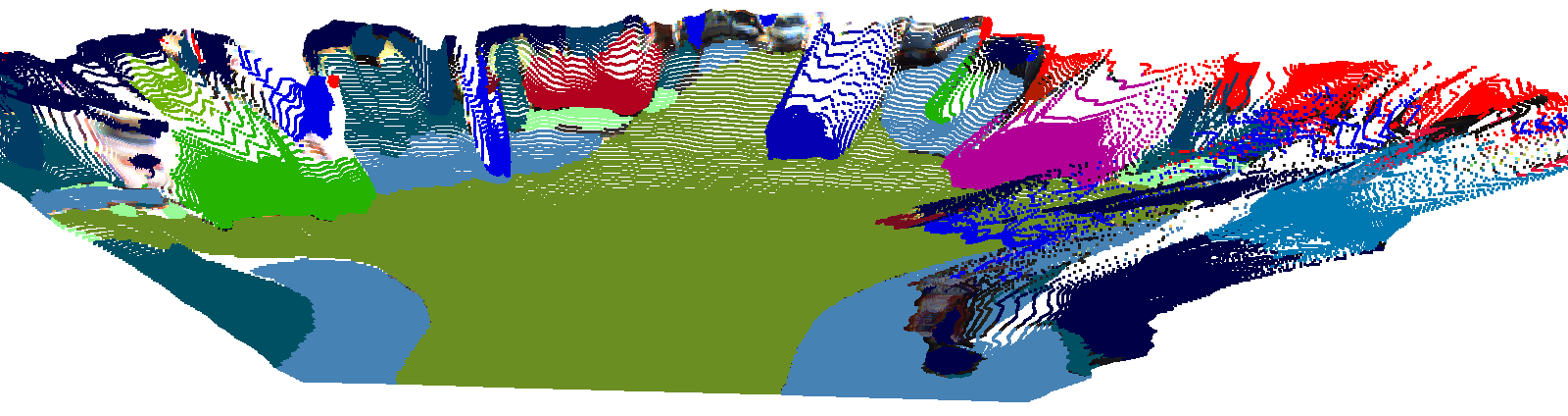}
        \\
        \includegraphics[width=0.15\linewidth]{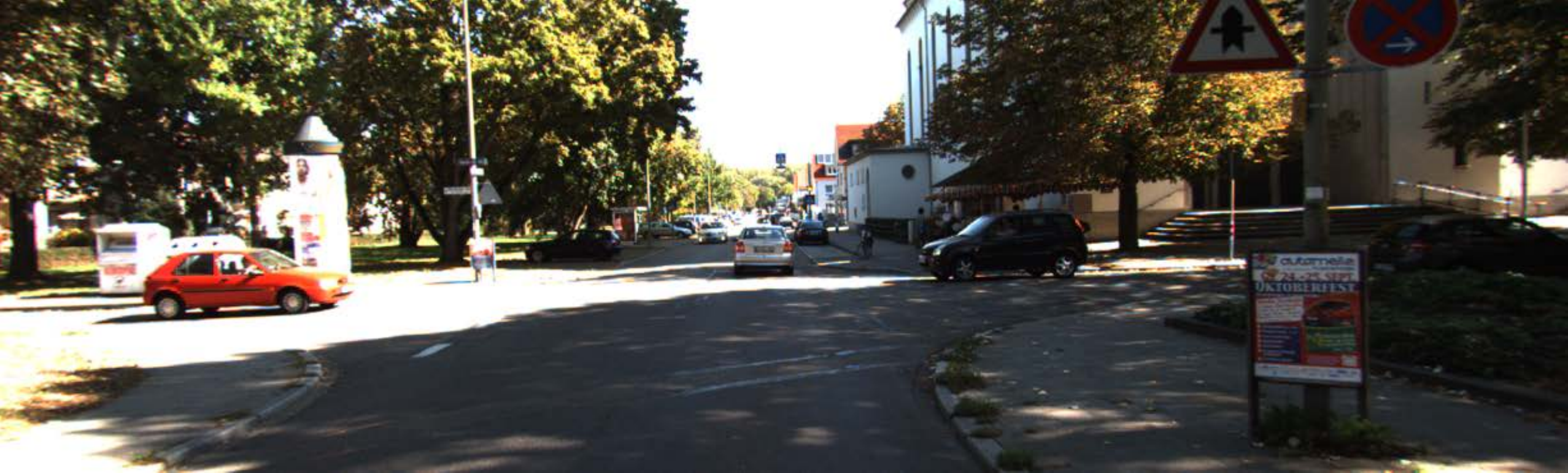}
        &
        \includegraphics[width=0.15\linewidth]{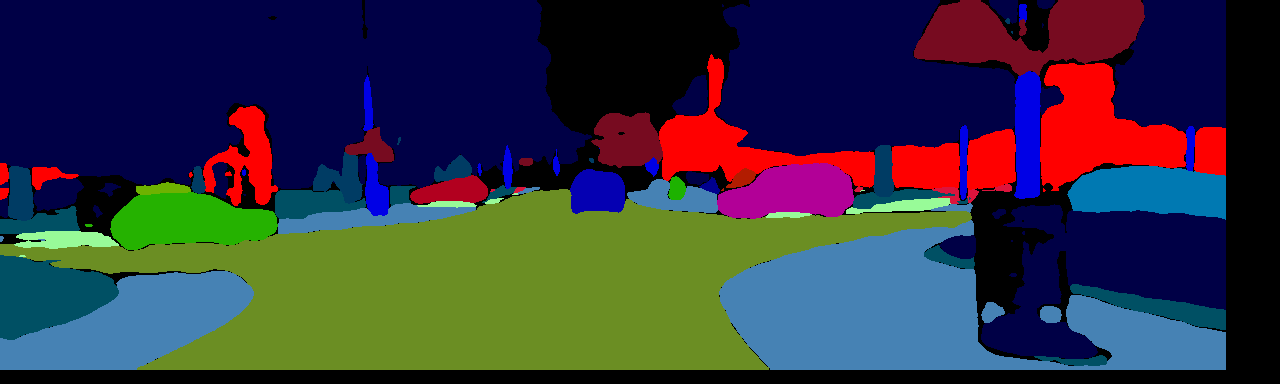}
        &
        \includegraphics[width=0.15\linewidth]{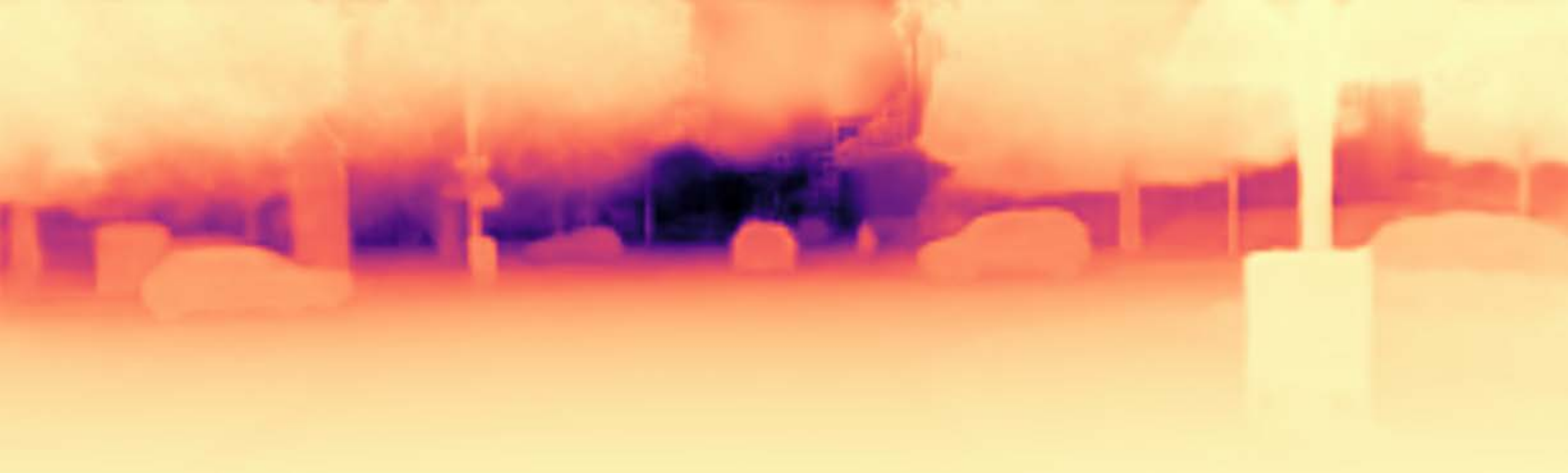}
        &
        \includegraphics[width=0.2\linewidth]{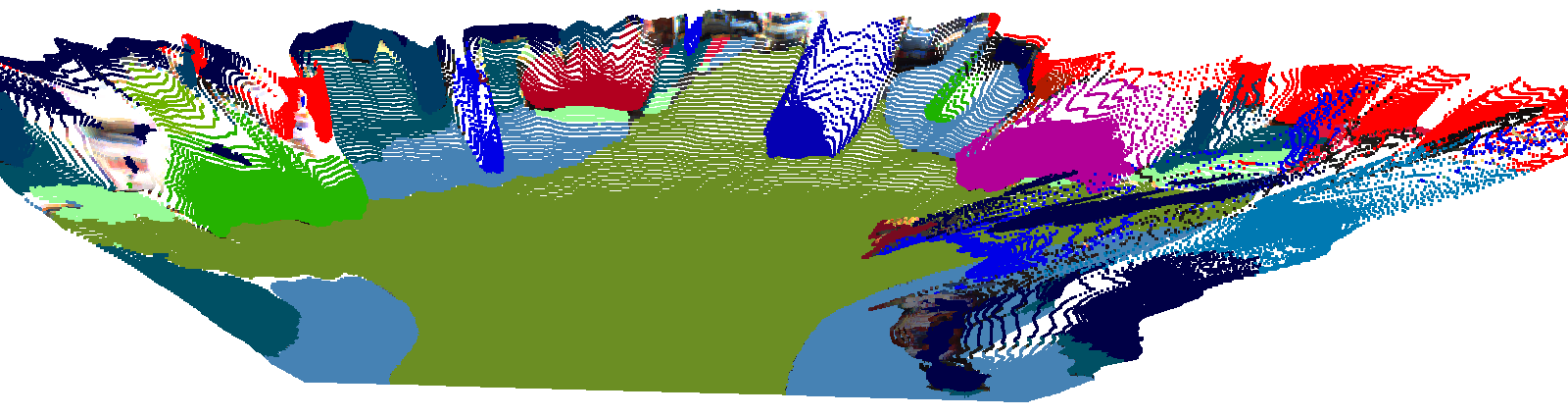}
        \\
        \includegraphics[width=0.15\linewidth]{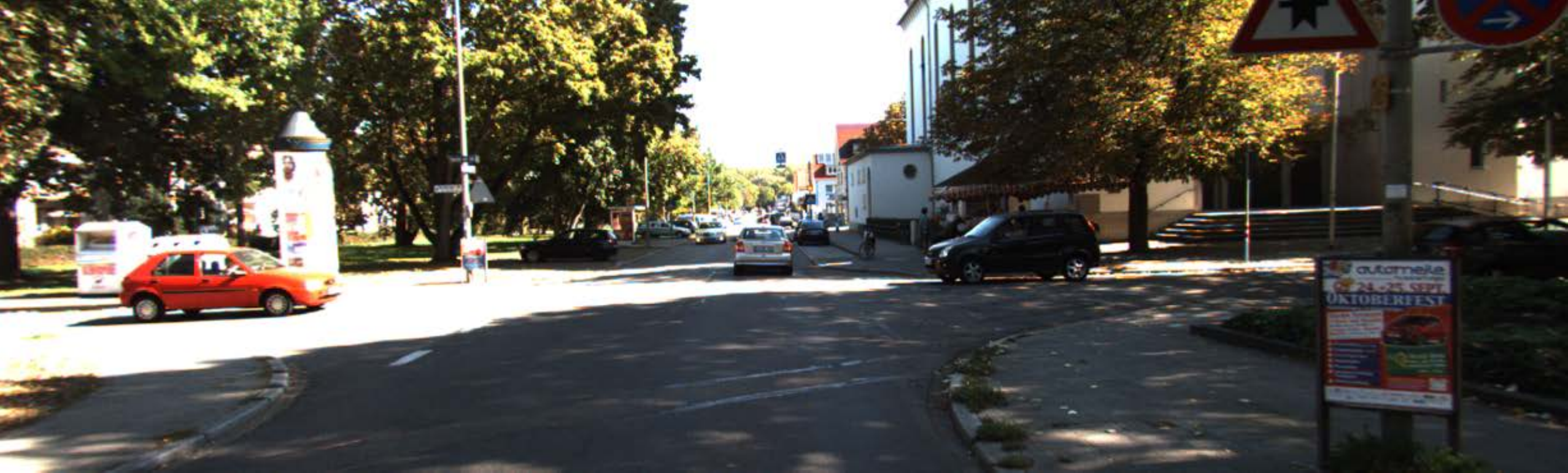}
        &
        \includegraphics[width=0.15\linewidth]{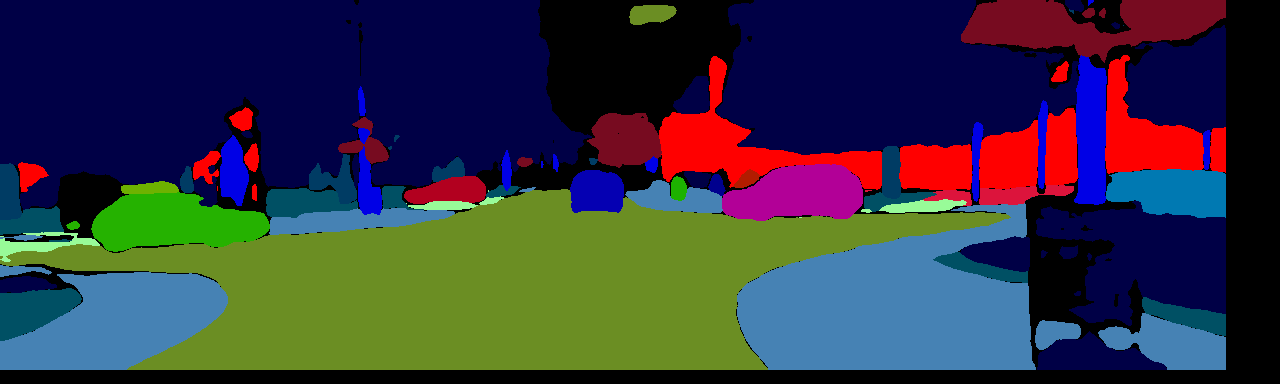}
        &
        \includegraphics[width=0.15\linewidth]{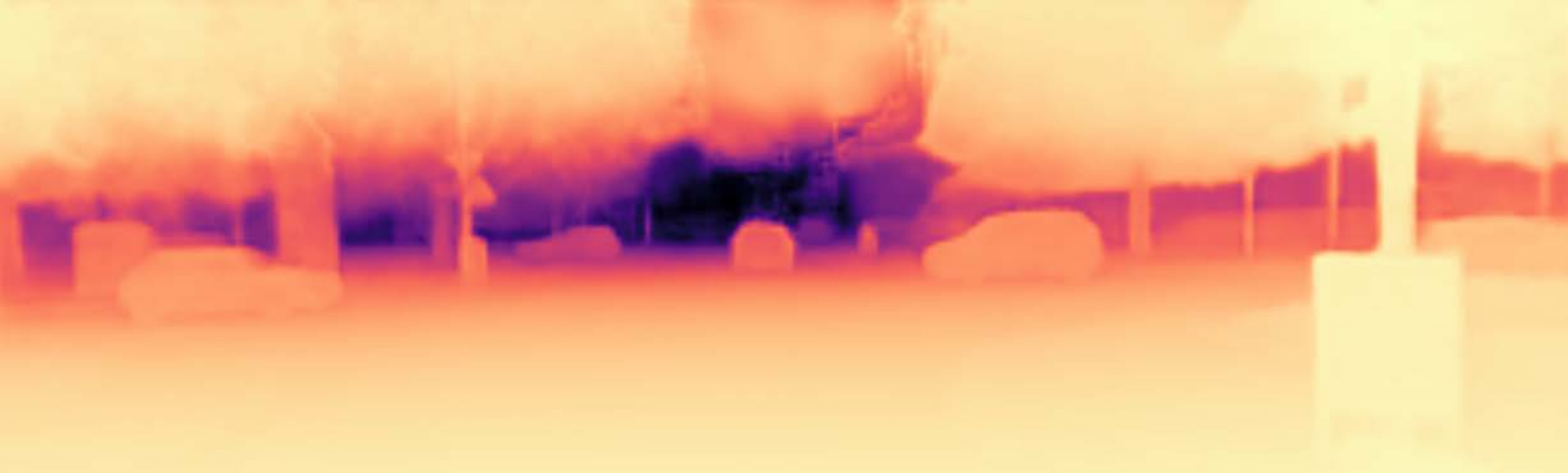}
        &
        \includegraphics[width=0.2\linewidth]{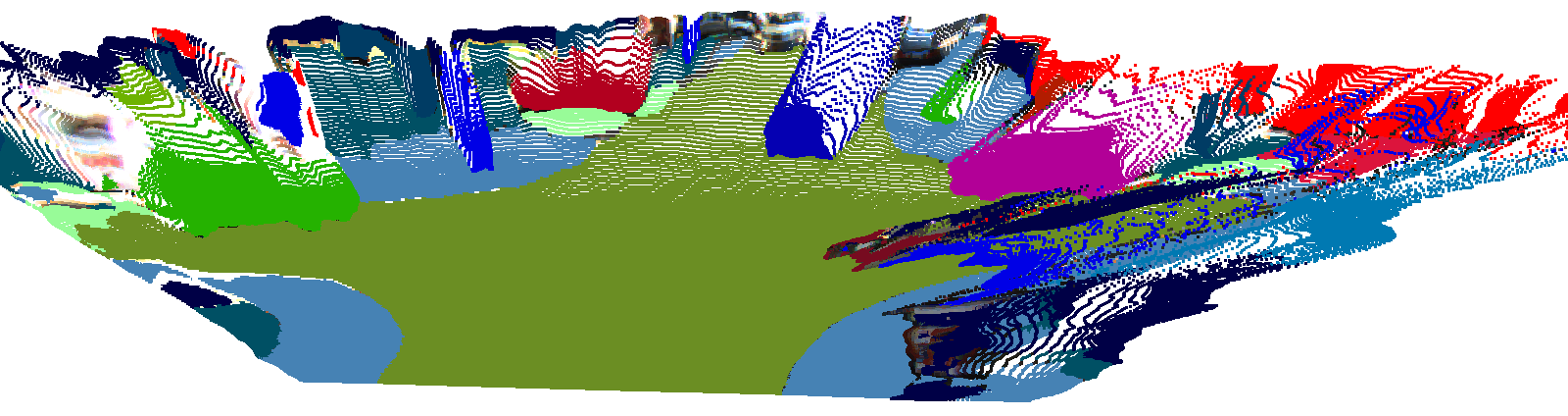}
        \\
        \includegraphics[width=0.15\linewidth]{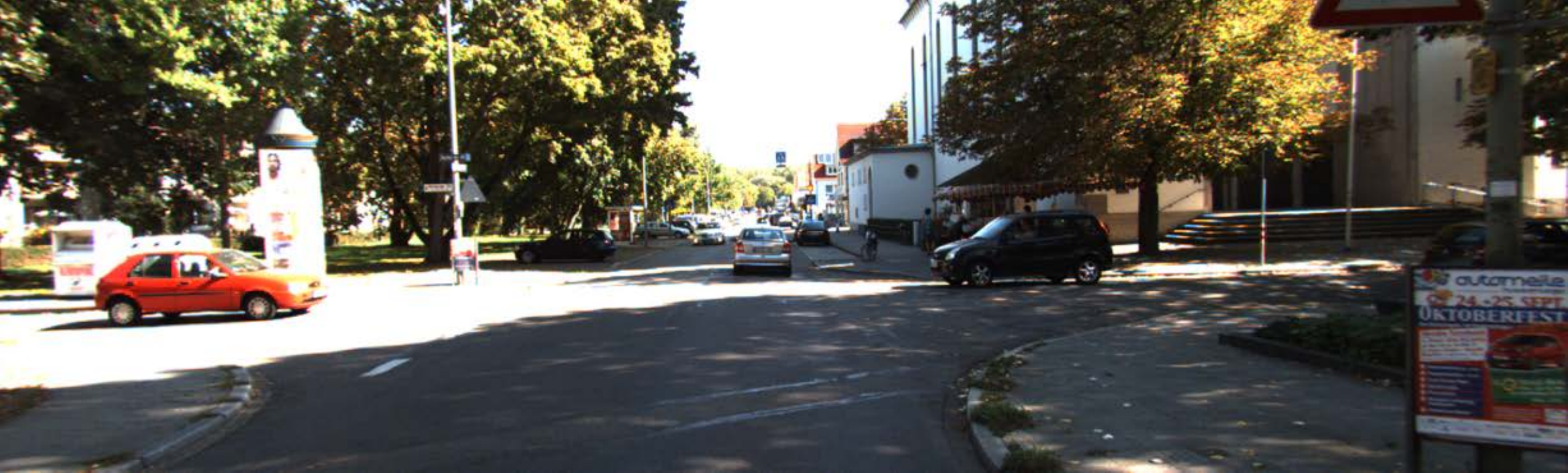}
        &
        \includegraphics[width=0.15\linewidth]{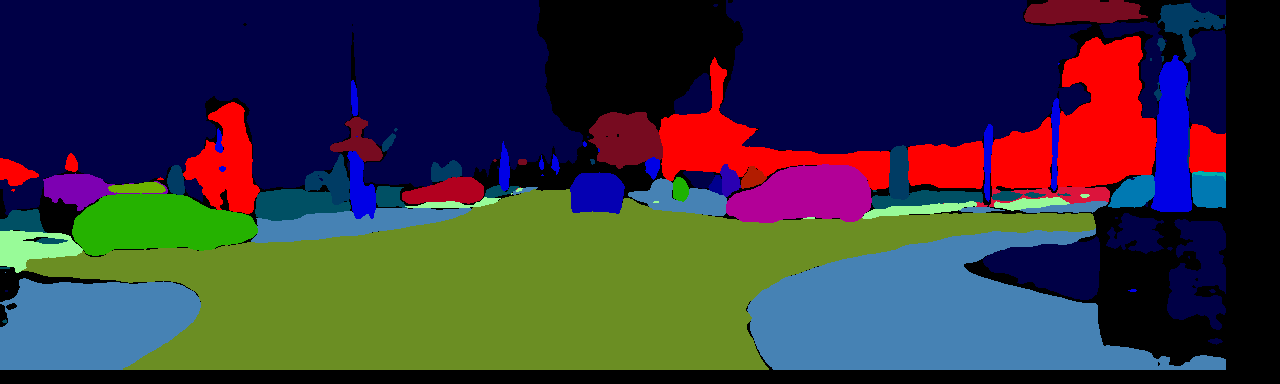}
        &
        \includegraphics[width=0.15\linewidth]{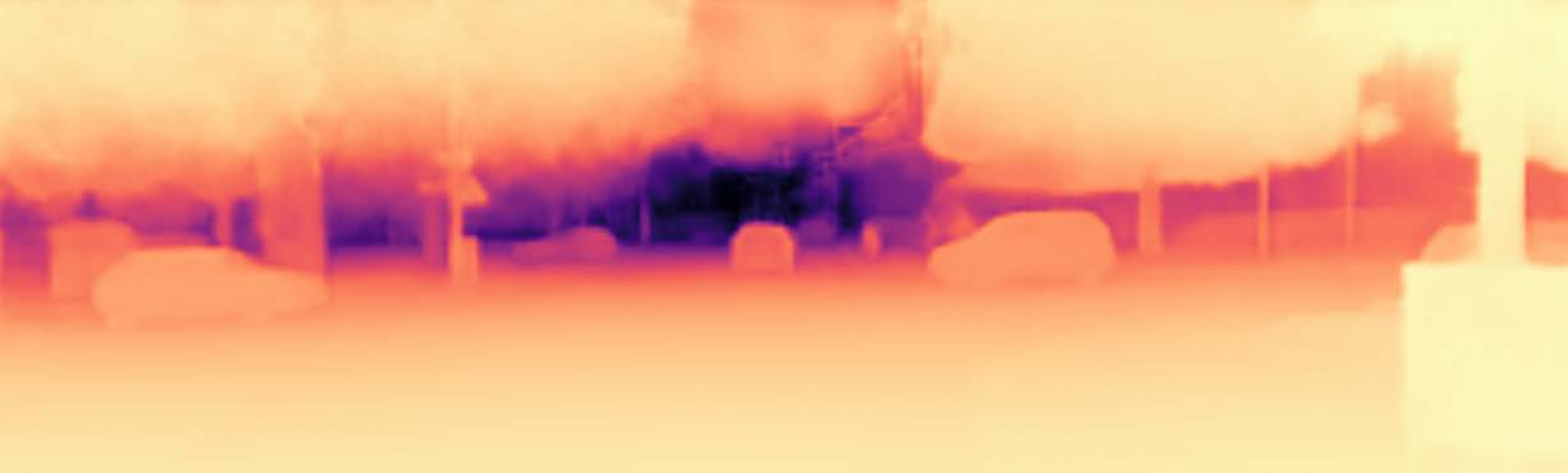}
        &
        \includegraphics[width=0.2\linewidth]{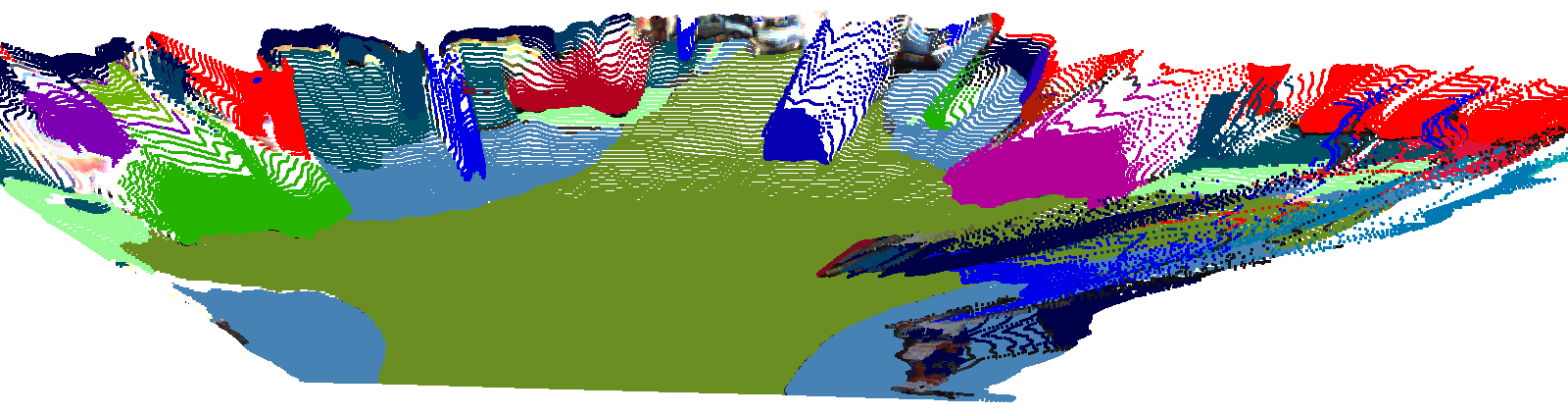}
        \\
        RGB & VPS & Depth & 3D \\
    \end{tabular}}
    \vspace{-5pt}
    \caption{
        \textbf{Qualitative results of \ourmodel}.
        The upper part is Cityscapes-DVPS, while the lower part is SemKITTI-DVPS.
        We show the sequence with input RGB images, video panoptic segmentation results (VPS), monocular metric depth estimation (Depth), and unprojected 3D prediction.
    }
    \vspace{-10pt}
    \label{fig:qualitative}
\end{figure*}

\subsection{Qualitative Results}
\label{sec:result:qualitative_results}
In \cref{fig:sota_comparison}, we present the qualitative comparison between \ourmodel and Uni-DVPS, which demonstrates the better segmentation and depth estimation quality.
We show more qualitative results on the Cityscapes-DVPS and SemKITTI-DVPS datasets in \cref{fig:qualitative}.
We plot the video panoptic segmentation in the image view, depicting the instance identity using a mask color.
We also provide the unprojected DVPS results as 3D visualization.

\section{Conclusion}
\label{sec:conclusion}
In this work, we introduce \ourmodel, an effective depth-aware video panoptic segmentation method that treats segmentation as the explicit scene representation process to estimate depth efficiently and associate objects in an online manner.
Our approach significantly outperforms existing SOTA methods on the established Cityscapes-DVPS and SemKITTI-DVPS benchmarks with fewer computational costs and higher inference speed, highlighting the effectiveness of our architecture in facilitating improved information flow and enhancing overall performance.
Through meticulous ablation studies, we systematically demonstrate the benefits of our explicit scene discretization design and the proposed online tracking with majority voting, leading to a holistic improvement of the DVPS framework in all its components.
The results underscore our work's pivotal contributions and the potential for safe autonomous driving.

{
    \small
    \bibliographystyle{ieeenat_fullname}
    \bibliography{main}
}

\end{document}